\newif\ifarxiv
\newif\ifralfinal
\ralfinalfalse \arxivtrue

\ifarxiv
\documentclass[letterpaper, 10 pt, conference]{ieeeconf}  
\fi
\ifralfinal
\documentclass[letterpaper, 10 pt, journal, twoside]{IEEEtran}  
\fi

\IEEEoverridecommandlockouts                              

\usepackage{graphicx}
\usepackage[table,xcdraw]{xcolor}
\usepackage{amsmath}
\usepackage{mathtools}
\usepackage{booktabs}
\usepackage{capt-of}
\usepackage{eso-pic}
\usepackage{hyperref}

\ifarxiv
\usepackage{fancyhdr}
\fi

\ifralfinal
\fi

\ifarxiv
\fi

\fancypagestyle{smallHeaders}{%
    \fancyhf{}
    \fancyhead[C]{ \fontsize{10}{10}\selectfont Preprint version; final version available at \url{http://doi.org/10.1109/LRA.2022.3146894}}
    \fancyfoot[C]{\thepage}
}%

\fancypagestyle{bigHeaders}{%
    \fancyhf{}
    \fancyhead[C]{ \fontsize{12}{12}\selectfont Preprint version; final version available at \url{http://doi.org/10.1109/LRA.2022.3146894}}
    
    \fancyfoot[C]{\thepage}
}%

\begin{document}

\title{ \ifarxiv\LARGE \bf\fi Improving Road Segmentation in Challenging Domains Using Similar Place Priors}

\author{Connor Malone, Sourav Garg, Ming Xu, Thierry Peynot and Michael Milford
\thanks{The authors are with the School of Electrical Engineering and Robotics, Queensland University of Technology, Brisbane 4000, Australia.}%
\thanks{Source code is publicly available at https://github.com/CMalone-Jupiter/SimilarPlaces.}
}

\maketitle
\ifarxiv
\thispagestyle{bigHeaders}
\pagestyle{smallHeaders}
\fi

\begin{abstract}

Road segmentation in challenging domains, such as night, snow or rain, is a difficult task. Most current approaches boost performance using fine-tuning, domain adaptation, style transfer, or by referencing previously acquired imagery. These approaches share one or more of three significant limitations: a reliance on large amounts of annotated training data that can be costly to obtain, both anticipation of and training data from the type of environmental conditions expected at inference time, and/or imagery captured from a previous visit to the location. In this research, we remove these restrictions by improving road segmentation based on \textit{similar} places. We use Visual Place Recognition (VPR) to find \textit{similar but geographically distinct} places, and fuse segmentations for query images and these similar place priors using a Bayesian approach and novel segmentation quality metric. Ablation studies show the need to re-evaluate notions of VPR utility for this task. We demonstrate the system achieving state-of-the-art road segmentation performance across multiple challenging condition scenarios including night time and snow, without requiring any prior training or previous access to the same geographical locations. Furthermore, we show that this method is network agnostic, improves multiple baseline techniques and is competitive against methods specialised for road prediction.
\end{abstract}

\section{Introduction}
\ifralfinal
\else
Scene
\fi understanding for mobile robots is crucial to enabling their safe operation, especially in the context of autonomous vehicles or systems that operate in populated environments. For a robot to understand the world around it, it must perform key tasks including traversability analysis, depth estimation, object detection, localisation, semantic segmentation~\cite{garg2020semantics} and more specifically for the focus of this paper, road segmentation. The semantic segmentation task in particular has benefited from the wide-spread adoption of modern neural networks; state-of-the-art systems can typically semantically segment scenes captured in \textit{ideal conditions} (i.e. clear atmosphere, well illuminated) to a high level of accuracy (e.g. mIoU $\geq$83\%~\cite{chen2018encoder} on the Cityscapes~\cite{cordts2016cityscapes} validation set). Accordingly, road segmentation in these ideal conditions can be considered a mostly solved task.

\newcommand{\schemscale}{0.17}
\begin{figure}
    \centering
    \includegraphics[scale=\schemscale,trim=0.8cm 1.2cm 0cm 0cm, clip=true]{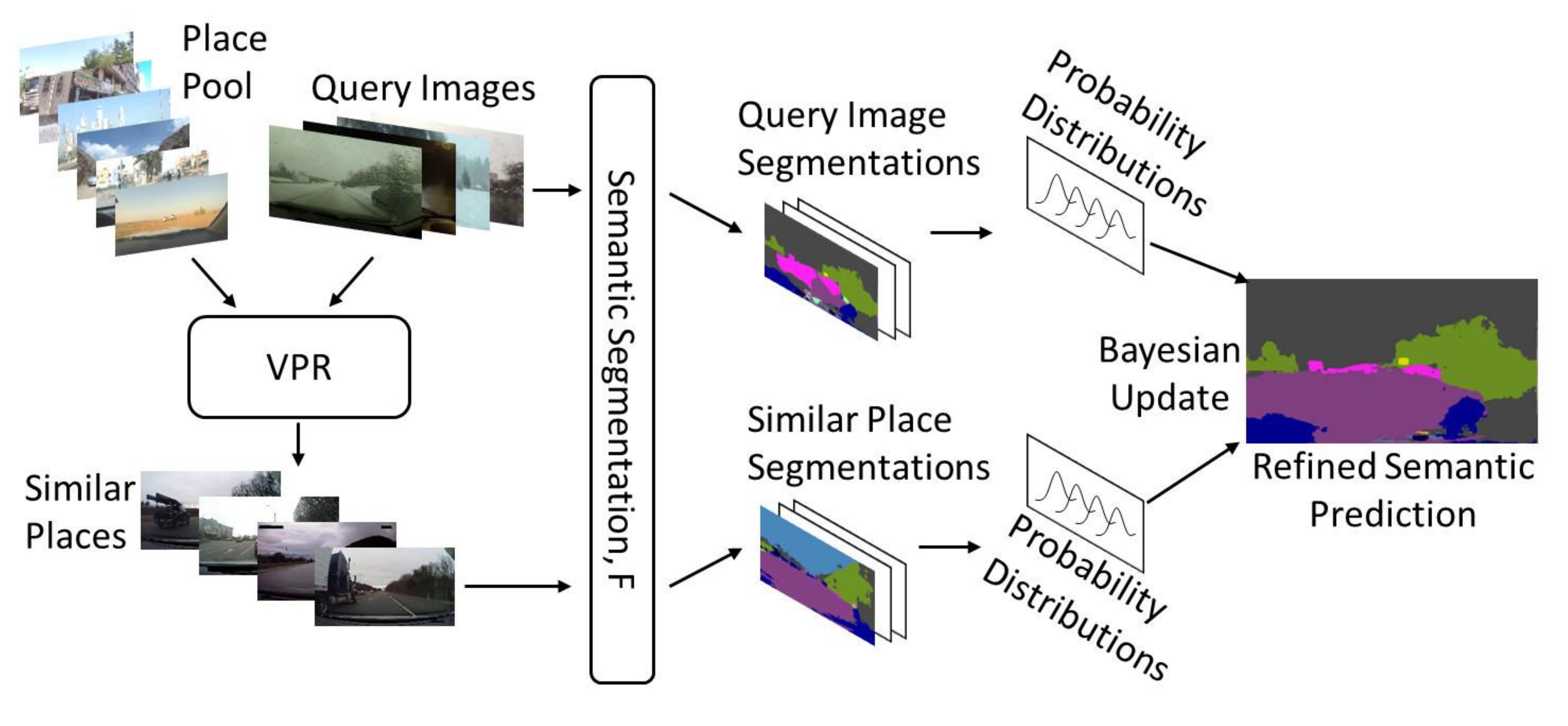}
    \caption{Road segmentation in challenging conditions such as at nighttime or in snow can be improved directly using priors from \textit{similar} places in more favourable conditions, without the need for previous access to the location, additional annotation, training or image manipulation.}
    \label{fig:Schem}
\end{figure}

If non-ideal conditions are both anticipated \textit{and} suitable exemplar data is available, approaches involving fine-tuning, domain adaptation or style transfer can significantly improve performance~\cite{sakaridis2020map, dai2018dark, porav2019don}. The viability of these approaches is limited however: beyond needing to anticipate the conditions ahead of time, they typically need large scale datasets and/or corresponding annotations to train the systems. Obtaining ground truth annotations can also be particularly difficult in challenging conditions, as evidenced by the need in past research to classify some pixels as `uncertain'~\cite{sakaridis2020map}. As robots move into ever more challenging and open-ended operational domains, predicting (and training for) the range of encountered conditions will become increasingly infeasible. Therefore it is important that novel methods are explored which complement existing approaches while removing some of these practical limitations.

Here we present an alternative approach that is based on three core observations: 1) that it's easier to perform semantic segmentation under ideal rather than challenging environmental conditions; 2) that many geographically distinct places in the world are similar, and 3) that VPR tends to find places with structural similarity, hence the focus on road segmentation for ground-based vehicles here. Together these insights form the foundation for our proposed \textit{similar place prior}-based approach to improving road segmentation, with the following contributions:

\begin{itemize}
\item A segmentation-network agnostic method based on a novel concept of finding \textit{similar} places to the query image, which improves road segmentation in challenging domains without any requirement for previous access to the actual location or the need for an extensive labelled data collection.
\item A Bayesian approach for fusing raw image segmentations captured under perceptually challenging conditions with segmentations from similar places captured under ideal conditions, retrieved through VPR. The fusion produces updated, improved class predictions by dynamically weighting the contribution from the similar place prior.
\item Extensive experimental evaluations across a range of environmental conditions (night, snow, rain) that demonstrate new levels of absolute road segmentation performance compared to state-of-the-art methods and a universal improvement of multiple state-of-the-art segmentation baselines using the proposed approach
\item Analysis revealing the effectiveness of current VPR for similar place retrieval and the specific performance benefits of the similar place segmentation process proposed here over a naive baseline implementation, and
\item New manually-curated subsets of WildDash 2 dataset for the segmentation task which explicitly separates performance evaluation across night, snow and rainy conditions, and shows that night still remains more challenging than snow and rain.
\end{itemize}

The paper proceeds as follows. We review semantic segmentation, visual place recognition and the use of priors in Section~\ref{RELATEDWORKS}. Section~\ref{METHOD} details the Bayesian update approach for incorporating segmentations. In Section~\ref{EXPERIMENTALSETUP}, we detail the benchmark datasets, experimental process and evaluation procedures. Results are presented in Section~\ref{RESULTS}, with discussion and future work possibilities in Section~\ref{DISCUSSIONANDFUTURE WORK}.

\section{Related Work}
\label{RELATEDWORKS}

Here we briefly touch on relevant research in semantic segmentation under challenging conditions and visual place recognition. 

\subsection{Semantic Segmentation}
Semantic segmentation involves the separation of an image into a set of analogous regions that can be categorised into a finite set of classes. This has been achieved through numerous techniques including colour space analysis, machine learning and most recently with great success, deep learning. With the development of fully convolutional networks for pixel-wise predictions~\cite{long2015fully}, deep learning has resulted in neural networks that are capable of highly accurate dense semantic label predictions on elementary datasets~\cite{chen2018encoder, girshick2015region, badrinarayanan2017segnet, zhao2017pyramid, tao2020hierarchical}, at least in ideal conditions. Achieving this level of accuracy in challenging domains, such as nighttime~\cite{sakaridis2020map,dai2018dark} or snow~\cite{lei2020semantic,vachmanus2020semantic}, remains a mostly unsolved research problem. Of these approaches, ~\cite{sakaridis2020map} is most conceptually similar to ours, except that our approach does not require any prior experience in the specific location.

Common approaches to improving segmentation performance in challenging domains include fine-tuning networks~\cite{li2020segmenting, valada2017adapnet}, sequential imagery~\cite{pfeuffer2020robust} and sensor fusion~\cite{pfeuffer2019robust}, most of which require either anticipation of the likely deployment conditions or the ability to collect training data representative of those conditions.
As an alternative approach, \cite{dai2018dark,sun2019see,romera2019bridging,sakaridis2020map} use domain adaptation to adapt networks trained on daytime datasets to the nighttime domain;~\cite{dai2018dark,sakaridis2020map} achieve this adaptation gradually through the use of intermediary twilight images. Related work on style transfer uses a similar concept to train generative adversarial networks (GANs~\cite{goodfellow2014generative}) to project an image from one domain into the style of another~\cite{zhu2017unpaired}. \cite{porav2019don} uses style transfer to train a set of domain adaptors that can be used to pre-process any image into a domain compatible with off-the-shelf segmentation networks. Our proposed approach shares some conceptual similarity around augmenting off-the-shelf segmentation networks, but instead achieves this goal using `similar place templates' instead of domain adaptation, offering several significant practical advantages.

Some existing approaches are similar to ours from the point of using prior segmentations to improve results, but typically have at least two significant extra requirements: the availability of images from the same geographical location and labelled images. Both~\cite{jonchery2018stixel} and \cite{schroeder2019using} have explored the use of priors for semantic segmentation but under relatively idealized conditions. \cite{schroeder2019using} improved segmentation performance on daytime images by fusing the results from temporal priors, while~\cite{jonchery2018stixel} used a probabilistic model and manually-labelled satellite images from the \textit{same} geographical locations to improve daytime segmentation. Similarly~\cite{sakaridis2020map} uses localised priors for domain adaptation.

\subsection{Visual Place Recognition}
Our proposed approach relies on the identification of \textit{similar} reference images using Visual Place Recognition systems~\cite{lowry2016visual}, warranting a brief overview. The key challenges of the VPR problem involve dealing with variations in camera viewpoint and scene appearance. From large-scale appearance-based mapping techniques~\cite{cummins2008fab} to appearance-robust day-night place recognition~\cite{milford2012visual}, recent VPR research focus has shifted more to the use of learning-based techniques to better represent and match places under a range of practical challenges~\cite{arandjelovic2016netvlad, chen2017deep, naseer2017semantics, garg19Semantic}. For a relatively constrained scenario, for example, where places are revisited with roughly a similar viewpoint, view-dependent image descriptors in particular demonstrate high performance, especially under varying scene appearance conditions~\cite{milford2012visual, naseer2017semantics, chen2017deep}. Although an ``apples to apples" comparison between VPR and semantic segmentation is not possible, VPR techniques can provide a ``robust'' means to look up similar images that can then provide segmentation priors, which result in performance improvements as presented in this paper.

\section{Method}
\label{METHOD}

We begin by defining some terminology referred to consistently within the remainder of the paper. A \textit{query image} refers to an image captured in a challenging domain (e.g. night, snow or rain). \textit{Reference images} refers to a set of images captured under ideal appearance conditions, but not from the test locations. We use VPR to retrieve \textit{similar} images from a place recognition perspective from a reference dataset and use these to construct an intermediate \textit{similar place template}. This similar place template is then used to update the prediction of the query image road segmentation. To maximize the generality of the method, we explicitly enforce that there be no exact geographic overlap between query and retrieved reference images, removing the need for any previous visit to the specific location. Finally, we refer to \textit{class logits} as the pre-softmax layer output of a given segmentation network.

We adopt a Bayesian approach to improving road segmentation in challenging domains, setting Gaussian priors over class logits for individual pixel/class combinations within a given query image. Prior distributions are parameterized using the similar place template, discussed in detail in Section~\ref{ssec:SimPR}. Notably, the uncertainty in our prior distributions is set using a dynamic weighting scheme. Priors are updated using the observed query class logits, yielding a posterior distribution. This posterior is then used to make the final class prediction. We only perform this Bayesian update for potential road pixels since we are addressing road segmentation specifically in this work. Figure~\ref{fig:Schem} summarizes our proposed methodology, which will be described in subsequent sections.

\subsection{Similar Place Template Construction}
\label{ssec:SimPR}

We now define the notion of \textit{similar places} for a query image, and use this to introduce our proposed method for constructing similar place templates. A similar place template is a $H\times W \times N_c$ tensor containing class logits where $H, W$ are the dimensions of the query image and $N_c$ is the number of semantic classes. The template is constructed by averaging class logits extracted from a subset of $k$ images from the reference set which are most similar to the query image.

We can identify these $k$ most similar places from the reference set by retrieving the best matches provided from a VPR method, choosing one with a tendency to find matches with similar structural and class-type layouts. For example, we expect that buildings, sidewalks and roads are in similar positions across query and retrieved images. Figure~\ref{fig:Schem} illustrates examples of places that are similar but not the same location. We provide a discussion around how we select relevant VPR methods to ensure that retrieved images exhibit this characteristic.

Note that our usage of VPR is distinct from the typical goal of VPR itself, which is to retrieve elements of the reference dataset corresponding to the \textit{same place}. We only use VPR as a tool for identifying places in the reference dataset which are \textit{similar}, to improve road segmentation performance. We now detail our method for constructing similar place templates.

\subsubsection{Similar Place Retrieval}
In the VPR literature, a variety of place representation methods have been developed that exhibit unique characteristics with regards to viewpoint and appearance-invariance. While viewpoint-invariant methods such as NetVLAD~\cite{arandjelovic2016netvlad} aggregate deep convolutional features without consideration of their spatial coordinates, viewpoint-aware methods like HybridNet~\cite{chen2017deep} directly match the full convolutional tensor. This property is desirable here because it leads to place matching based on retrieving images with similar scene layout for a given query. Given our desired characteristics of similar places~i.e. those that are structurally and semantically aligned, HybridNet is a strong candidate for our underlying VPR method. We use the last convolutional layer, specifically its feature tensor to represent places. Similar place retrieval is then performed through identifying reference images with lowest Cosine distance feature tensors to the query image.

Let $x_{ij, n}^s$ represent the value of the similar place template at the $i, j$ pixel coordinate for class $n$. We compute this by taking the mean of the class logits at the same pixel coordinate and class across the $k$ retrieved reference images, where $k\geq1$. Furthermore, let $\sigma^s_n$ be the class-wise sample standard deviation of similar place template values over all pixels where class $n$ has the largest class logit value (hence largest class prediction probabilities).

\subsection{Bayesian Update For Segmentation Logits}

We now introduce our Bayesian update in detail, first by specifying the prior distribution and how it is parameterized by similar place templates, the likelihood and finally the update step over semantic class logits.

\subsubsection{Prior Distribution}\label{sssec:prior}

Let $x_{ij, n}$ be the true unknown class logit for class $n$ at pixel coordinate $i,j$ for the query image. We specify a Gaussian prior over this logit value, $x_{ij, n}\sim \mathcal{N}(x_{ij, n}^s, (\omega_n \sigma^s_n)^2)$, parameterized by the similar place template values and class-wise (identical for all pixels) standard deviations. We can decompose the prior standard deviation into observed variability for template logits $\sigma^s_n$ and a dynamic tempering factor $\omega_n$ which further modulates the confidence in the prior based on class coverage, which will be discussed in Section~\ref{ssec:weights}. 

\subsubsection{Likelihood and Bayes update}

We specify a Gaussian likelihood for an observed logit for class $n$ from the query segmentation model $x_{ij, n}^q$ given true unknown logit $x_{ij, n}$; this ensures the posterior is also Gaussian. Concretely, $x_{ij, n}^q \sim \mathcal{N}(x_{ij, n}^s, {\sigma^{q}_n}^2)$, where $\sigma^q_n$ is computed identically to $\sigma^s_n$ but applied to the query image. The posterior follows as
\begin{equation}\label{eq:posterior}
    x_{ij, n} \sim \mathcal{N}(x_{ij, n}^*, {\sigma_n^*}^2),
\end{equation}
where the posterior mean and variance $x_{ij, n}^*$, ${\sigma_n^*}^2$ are
\begin{equation}\label{eq:posteriorparams}
    {\sigma_n^*}^{2} = \frac{1}{{\sigma^{q}_n}^2 + ({\omega_n\sigma_{n}^s})^{2}}, \quad x_{ij, n}^* = {\sigma_n^*}^{2}\left(\frac{x_{ij,n}^q}{{\sigma^{q}_n}^{2}} + \frac{x_{ij,n}^s}{\omega_n^2{\sigma_{n}^s}^{2}}\right)
\end{equation}
For each pixel, taking the maximum \textit{posterior} mean across all classes will yield the new class prediction.

\subsection{Prior Rescaling Factor}\label{ssec:weights}

We now introduce our novel dynamic tempering factor $\omega_{n}$ introduced in Section~\ref{sssec:prior}, which further modulates the variance of our prior distribution. We determine $\omega_n$ directly for class $n$ using the class logits in both the reference image set and query image. We apply $\omega_n$ on top of the observed class-wise logit standard deviation because we observed that this metric strongly correlates with segmentation quality. Better query segmentations typically result in larger values of $\omega_n$, putting more confidence in the query predictions for class $n$ with the converse being the case for retrieved reference images. Previously, comparison of Maximum Softmax Probabilities (MSP) has been utilised for evaluating segmentation quality ~\cite{hendrycks17baseline}. However, neural networks tend to produce high MSP values, making this an unreliable quality measure~\cite{xia2020synthesize}. Our proposed method is novel because it exploits the expected structure of an environment retrieved using VPR to determine the quality of a segmentation. 

Our tempering factor is computed using the class coverage percentages in query and reference images. Specifically, $\mathcal{C}_{n,q}$ represent the proportion of pixels in the query image where class $n$ has the highest class probability, whereas $\mathcal{C}_{n,k}$ and $\mathcal{C}_{n,\ell}$ correspond to the class coverage percentages across the set of $k$ and $\ell > k$ most similar retrieved reference images, respectively. The $\mathcal{C}_{n,\ell}$ term represents the class coverage over a larger set of similar places in the reference set, whereas $\mathcal{C}_{n,k}$ represents the class coverage of the similar place template. Define
\begin{equation}\label{eq:weight}
    \omega_{n} = \frac{|\mathcal{C}_{n,\ell} - \mathcal{C}_{n,k}|}{|\mathcal{C}_{n,q} - \mathcal{C}_{n,\ell}|}.
\end{equation}
$\omega_n$ decreases with the class coverage inconsistency of the query image to the larger set of similar places in the reference set, leading to more confidence in the prior. Conversely, $\omega_n$ increases with the inconsistency of the class coverage of the similar place template to the larger set of similar places, leading to less confidence in the prior.

\subsection{Identifying Potential Road Pixels}\label{ssec:RoadPix}

We isolate the updated class predictions to only pixels with the potential to be road (predictions and performance for other classes largely remains the same as for the query). These pixels are identified by taking the union of areas where the query and similar place template predict the road class. Pixels in this area are given the posterior class logits and others remain unchanged.

\section{Experimental Setup}
\label{EXPERIMENTALSETUP}

We evaluated the effectiveness of our proposed method using multiple benchmark datasets and semantic segmentation methods. In all experiments, we show the utility of our method as a \textit{post-processing} method to refine a base segmentation output. We first describe our datasets, followed by the state-of-the-art segmentation methods used.

\subsection{Datasets}

We used three benchmark datasets for evaluation which collectively include 1) test imagery captured under a variety of conditions that are challenging for segmentation networks and 2) reference images captured in ideal appearance conditions, but \textit{without} geographical overlap with the test images.

\paragraph{Dark Zurich~\cite{SDV19}} This dataset consists of $8779$ GPS-tagged images with fine pixel-wise annotations, captured at nighttime, twilight and daytime. For evaluation, it contains $201$ images ($151$ test \& $50$ validation) with $19$ evaluation classes used in the Cityscapes dataset~\cite{cordts2016cityscapes}. We evaluated our method across the 151 test images and utilised the $3041$ daytime training images as our reference set.

\paragraph{WildDash 2 ~\cite{Zendel_2018_ECCV}} The WildDash 2 dataset provides 4256 non-sequential public images covering a large global footprint and numerous weather conditions (sunny, snowy, rainy, flooding), times of day (including night) and different environments (urban, rural, off-road), with a range of camera characteristics (ideal, noisy, compressed, distorted). Ground truth semantic labels are provided in the commonly adopted Cityscapes format~\cite{cordts2016cityscapes}. The dataset was separated into 3265 daytime reference images, 364 nighttime, 247 rainy, 237 snowy and 143 `other' images. The splits details are publicly available.

\paragraph{Berkeley Deep Drive 100K ~\cite{bdd100k}} Berkeley Deep Drive 100K (BDD100K) contains 100k road scene images captured within America under a diverse range of environmental and weather changes along with GPS positions. We used a subset provided with the BDD100K dataset containing 10k images - BDD10K (7k training, 1k validation and 2k testing). We separated the training set into 6519 daytime reference images, 314 nighttime query images and 167 `other' images. A list of images in each split and code for separating the dataset is publicly available.

\subsection{Experiments}
We conducted a battery of experiments across the datasets to evaluate the effect on road segmentation performance of the proposed system. In particular, we evaluated the system using a range of off-the-shelf semantic segmentation networks and an array of VPR methods for similar place retrieval (Section~\ref{ssec:SimPR}).

\subsubsection{Semantic Segmentation Networks}
In our work we used three segmentation networks to provide baselines for our system to update and for performance comparisons. We chose two standard networks, DeepLabV3+\footnote{https://github.com/VainF/DeepLabV3Plus-Pytorch}~\cite{chen2018encoder} and RefineNet~\cite{Lin:2017:RefineNet}, with pre-trained models that report mean IOU scores of $72.1\%^1$ and $73.6\%$ on the Cityscapes test set~\cite{cordts2016cityscapes} respectively. In addition we evaluate MGCDA~\cite{sakaridis2020map}, a network trained for improved nighttime performance, for a comparison to state-of-the-art domain adaptation approaches. It uses RefineNet as a base network and was shown to perform better than several other domain adaptation systems (AdaptSegNet~\cite{tsai2018learning}, ADVENT~\cite{vu2019advent}, BDL~\cite{li2019bidirectional} and DMAda~\cite{dai2018dark}) on the Dark Zurich test data.

\subsubsection{Prior Selection}
We performed extensive work to determine the suitability of current VPR systems ~\cite{milford2012visual, arandjelovic2016netvlad, chen2017deep, naseer2017semantics, garg19Semantic} for this novel task of selecting a \textit{non-local} prior with high semantic similarity to the query. We chose to incorporate pre-trained implementations of NetVLAD~\cite{arandjelovic2016netvlad} and HybridNet~\cite{chen2017deep} due to their viewpoint-robustness properties. These systems allowed us to investigate the effects of viewpoint-dependence using features from different layers of HybridNet and viewpoint-invariance using NetVLAD, presented in Section~\ref{sec:VPRAblate}. For each query dataset in Table~\ref{tab:singleTable}, prior retrieval was made from a reference set containing the `ideal'/daytime images from the respective parent dataset. Where geographic separation could not be ensured by GPS, it was confirmed by manually inspecting retrieved images.

\begin{table}
\caption{Road IOU results comparing state-of-the-art baselines before and after applying our method.}
\centering
\resizebox{0.49\textwidth}{!}{
\begin{tabular}{@{}lccccccc}
\toprule
& Dark Zurich  & \multicolumn{3}{c}{Wild Dash 2}        &     BDD & All \\

\cmidrule(lr{0.75em}){2-2}
\cmidrule(lr{0.75em}){3-5}
\cmidrule(lr{0.75em}){6-6}
\cmidrule(lr{0.75em}){7-7}
\textbf{Method/Dataset}  & Night (Test)  & Night       & Snow        & Rain        & Night & Average       \\
\midrule
DeepLabv3 (DLv3)         & 71.0          & 49             & 54.7          & 50.8          & 32.02 & 51.5          \\
  + Dataset-Avg Prior           & 69.7          & 55.8          & 61.1           & 59.8          & 62.0 & 61.7         \\
  + \textbf{Ours}: Similar Place Prior          & 71.9          & \underline{60.8}          & 63.3           & \underline{63.1}          & \underline{67.3} & \underline{65.3}         \\
  + \textbf{Ours}: Prior+Query           & \underline{73.5}          & 58.2          & \underline{64.0}           & 62.4          & 65.0 & 64.6         \\
\midrule
RefineNet (RN)           & 68.9          & 46.9          & 61.8           & 66.9          & 70.0 & 62.9          \\
+ Dataset-Avg Prior          & 70.5              & 55.7          & 60.6          & 59.8          & 61.6 & 61.6           \\
+ \textbf{Ours}: Similar Place Prior    & 73.3              & \underline{60.7} & 66.3          & 65.5          & 69.7 & 67.1          \\
+ \textbf{Ours}: Prior+Query  & \underline{76.8} & 58.7 & \underline{70.6} & \underline{\textbf{70.8}} & \underline{75.2} & \underline{70.4} \\
\midrule
MGCDA (M)                   & \underline{\textbf{80.3}}  & 65.6          & 65.0  & 69.3  & \underline{\textbf{78.7}} & 71.8          \\
+ Dataset-Avg Prior           & 70.5          & 55.4          & 59.5          & 59.6          & 61.4 & 61.3         \\
+ \textbf{Ours}: Similar Place Prior           & 73.9          & 60.5          & 65.5           & 64.7          & 69.5 & 66.8         \\
+ \textbf{Ours}: Prior+Query           & 79.9          & \underline{\textbf{67.7}}          & \underline{\textbf{71.6}}           & \underline{70.7}          & 77.7 & \underline{\textbf{73.5}}         \\
\midrule
YOLOP           & 80.0          & 56.8          &  63.5           & 61.3          & 66.3    & 65.6       \\
\bottomrule
\end{tabular}
}
\label{tab:singleTable}

\end{table}

\begin{figure}
\centering
    \begin{tabular}{cc}
    \includegraphics[scale=0.1]{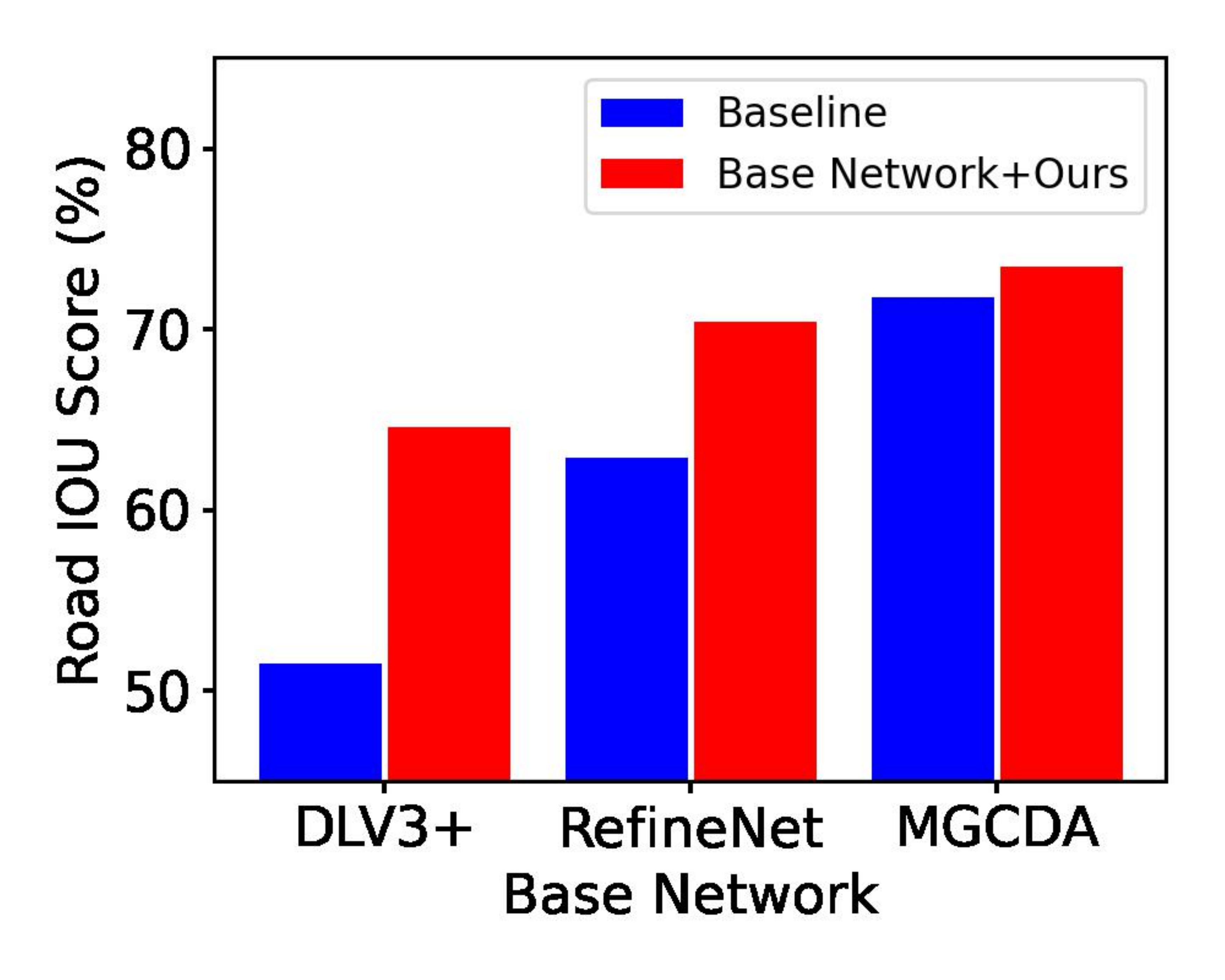} & 
    \includegraphics[scale=0.1]{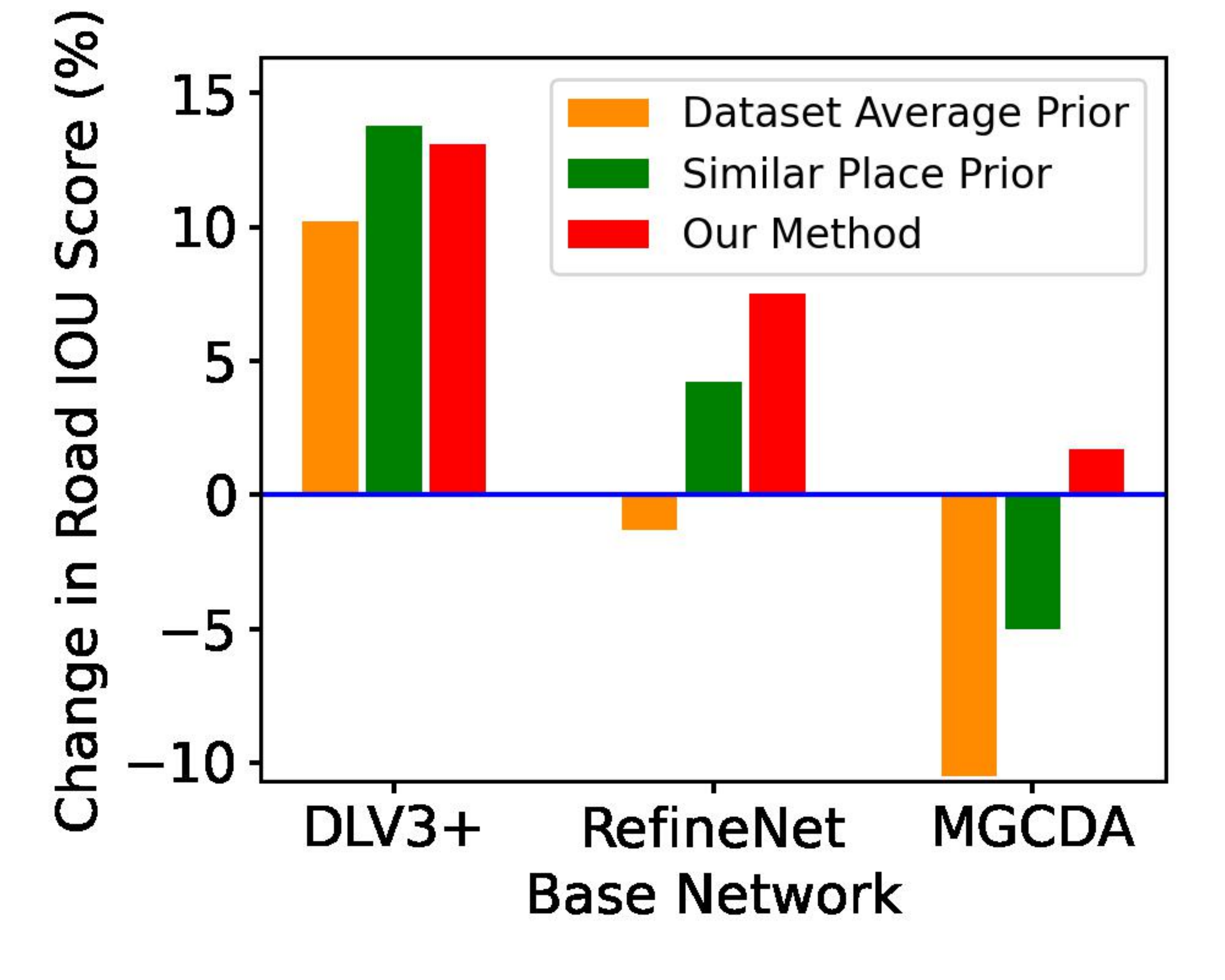} \\
    
    (a) & (b)
    \end{tabular}
    \captionof{figure}{Average road IOU across all datasets, showing our proposed method makes a universal improvement on all tested base networks.}
    \label{fig:TableSum}

\end{figure}

\subsection{Evaluation Metrics}
The standard evaluation metric for semantic segmentation is the Intersection over Union (IoU) score. This score computes the ratio between the area of intersection and area of union for the predicted and ground truth class labels. Specifically for our application, we computed IOU for the road class. As the baseline networks were trained only to predict the 19 Cityscapes evaluation classes, image regions in the WildDash 2 dataset that utilised additional labels were treated as undefined regions.

\section{Results}
\label{RESULTS}

In this section, we present road segmentation results highlighting the performance benefits achieved through applying our proposed method to the output of multiple state-of-the-art segmentation methods. Table \ref{tab:singleTable} and Figures \ref{fig:TableSum} and \ref{fig:QualResDL} demonstrate these results. We also considered other selected semantic classes based on an analysis of class-wise average softmax probabilities to observe the applicability of our method to different classes such as sky where mIoU (averaged over datasets) for RefineNet improved from 45.8 to 53.0. We present a series of ablations relating to the effect of choosing a VPR method, the number of candidates required for dynamic weighting, the behaviour of our method with varying baseline performance, the utility of ground truth labels and a comparison with specialised road segmentation networks. For the semantic segmentation networks used, no additional training was performed and furthermore, retrieved reference images have no direct geographical overlap with the query images. The convention for the data will follow that, unless stated otherwise, all similar places have been found using HybridNet's last convolutional layer ($conv6$) features and predictions have been updated using $k=5$ and $\ell=10$ in our system.

\newcommand{\scaleSGOne}{0.125\textwidth}

\begin{figure*}
    \centering
    \begin{tabular}{cccccc}
    Query Image & RefineNet & \textbf{RefineNet+Ours} & GT & MGCDA & \textbf{MGCDA+Ours} \\
    \includegraphics[width=\scaleSGOne]{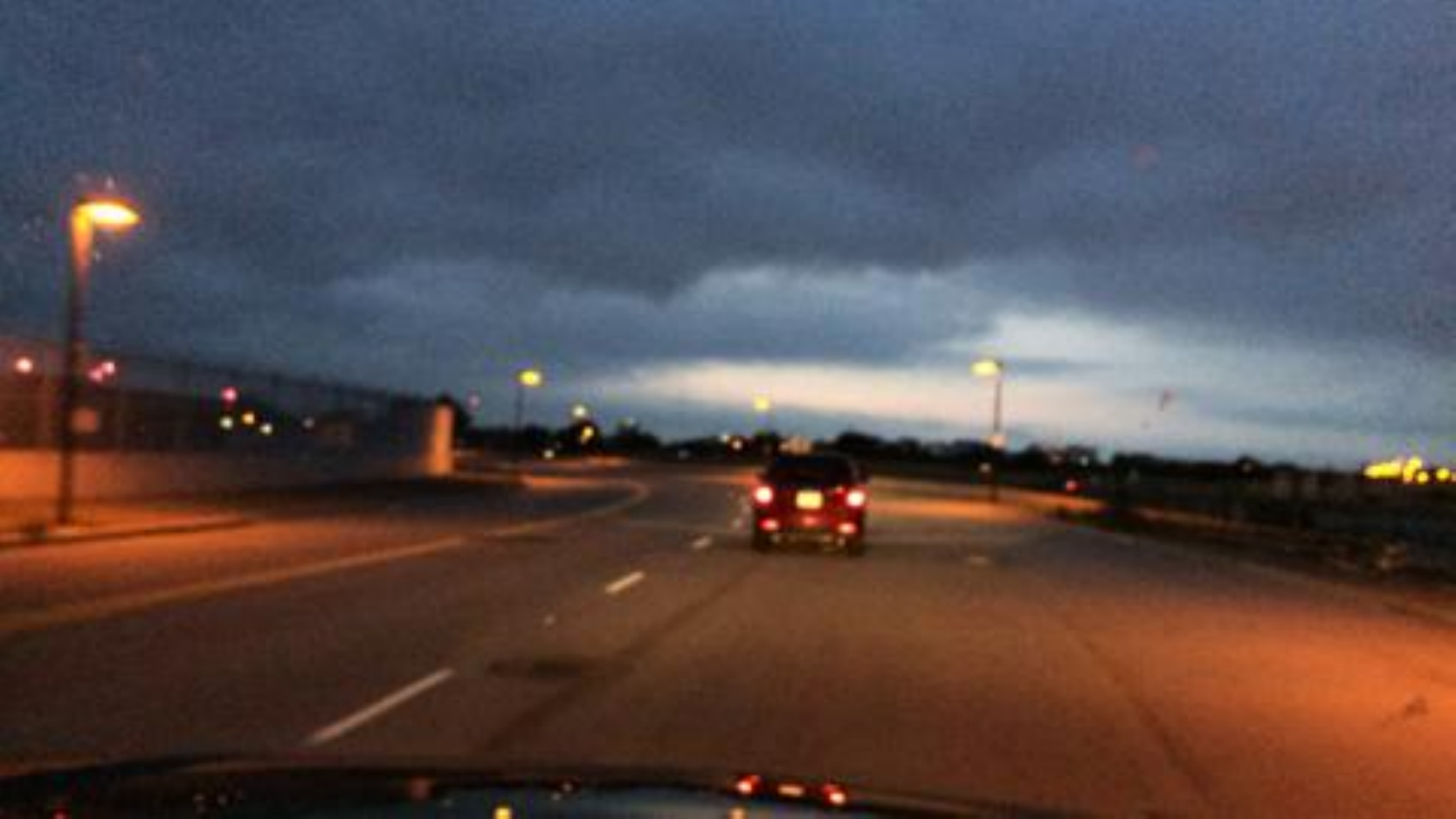} &
    \includegraphics[width=\scaleSGOne]{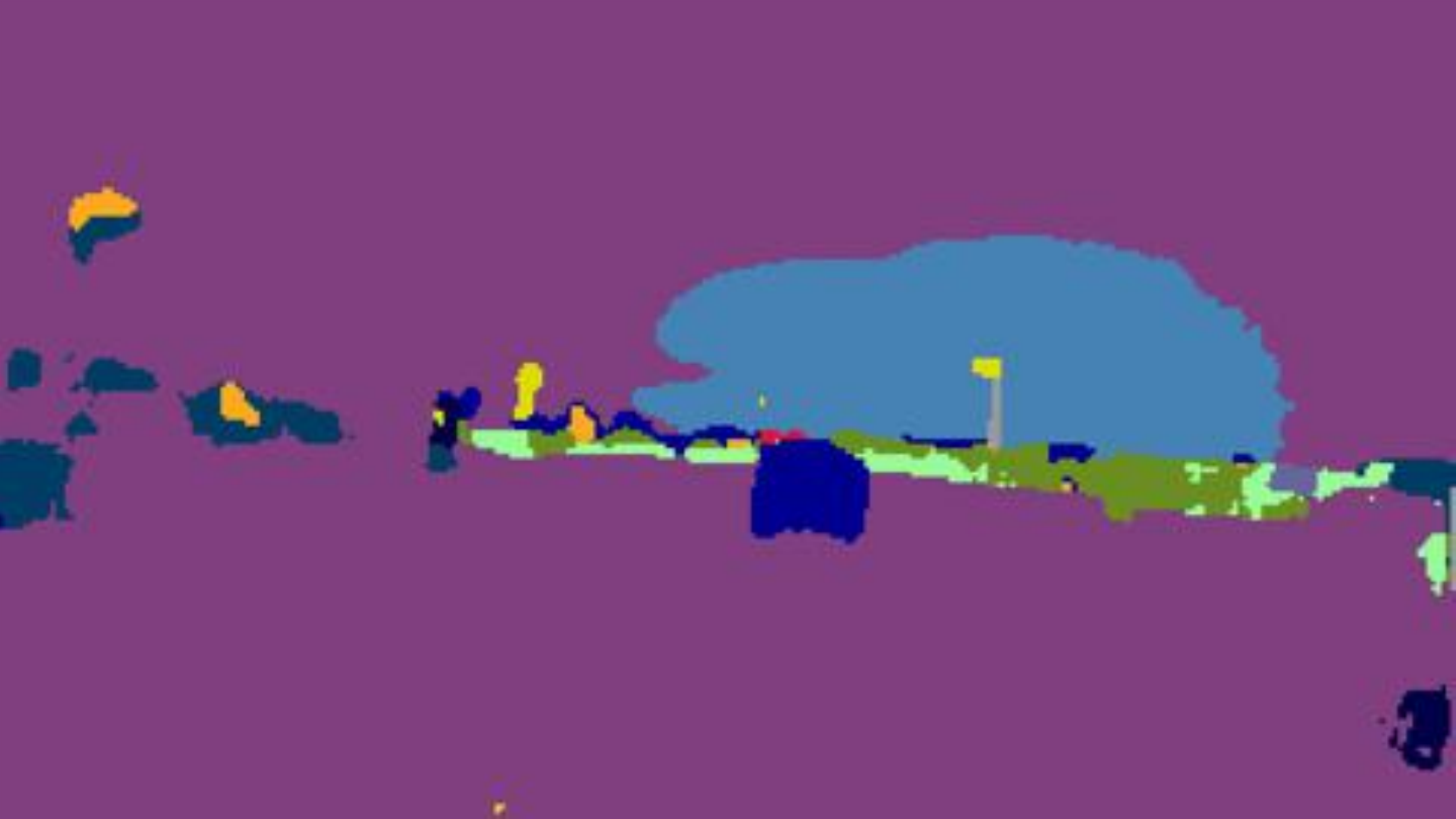} &
    \includegraphics[width=\scaleSGOne]{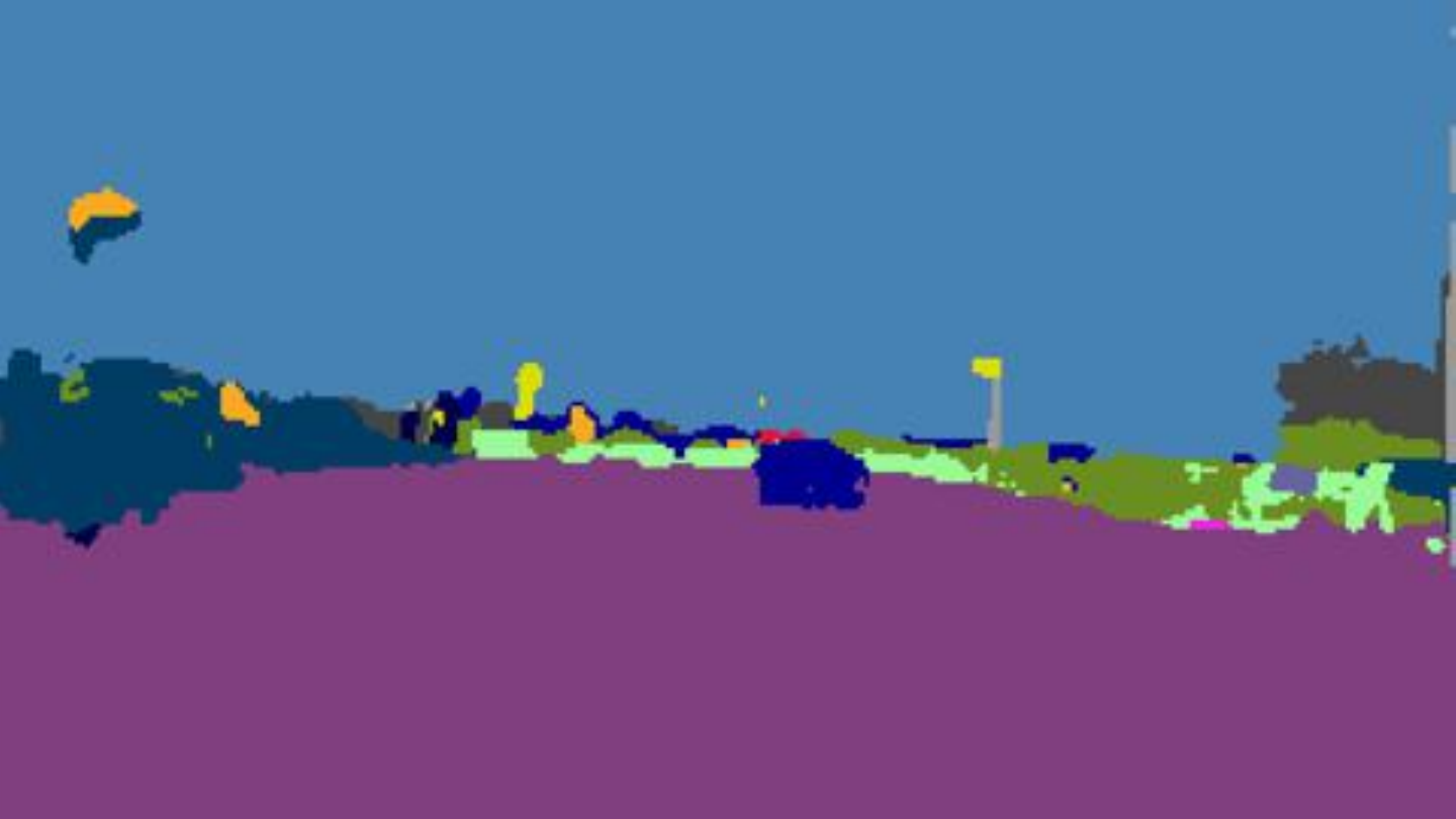} &
    \includegraphics[width=\scaleSGOne]{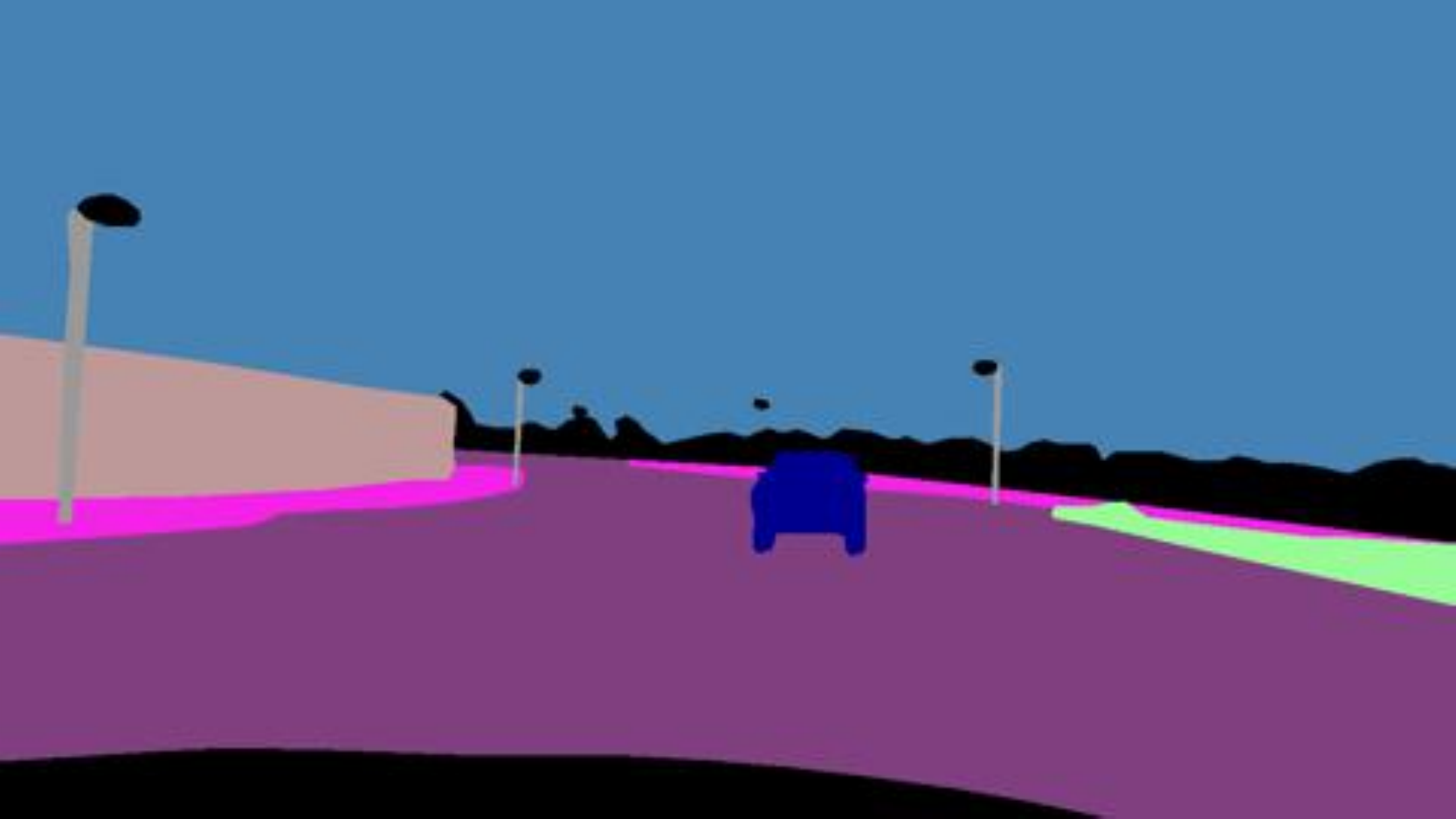} &
    \includegraphics[width=\scaleSGOne]{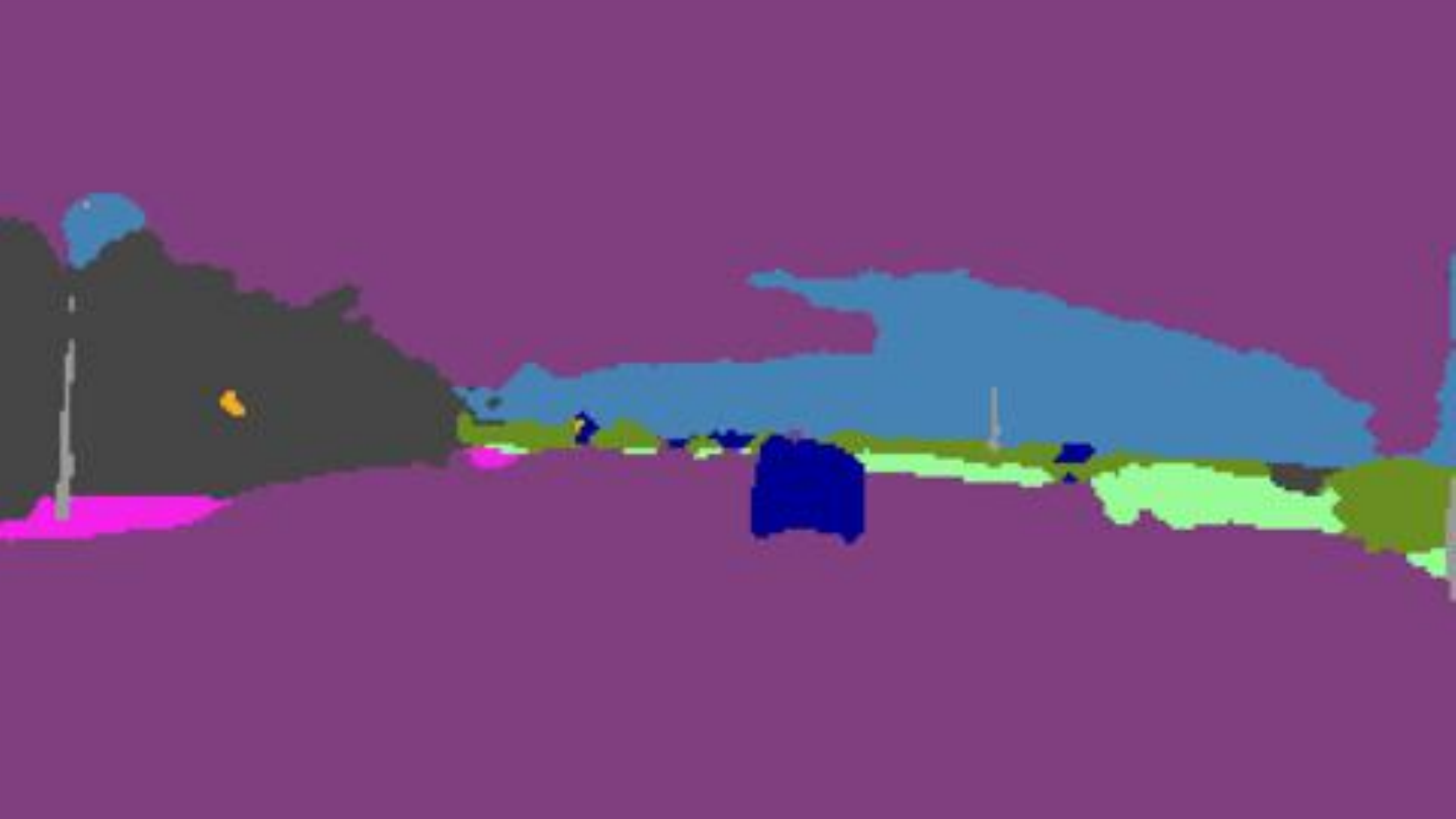} &
    \includegraphics[width=\scaleSGOne]{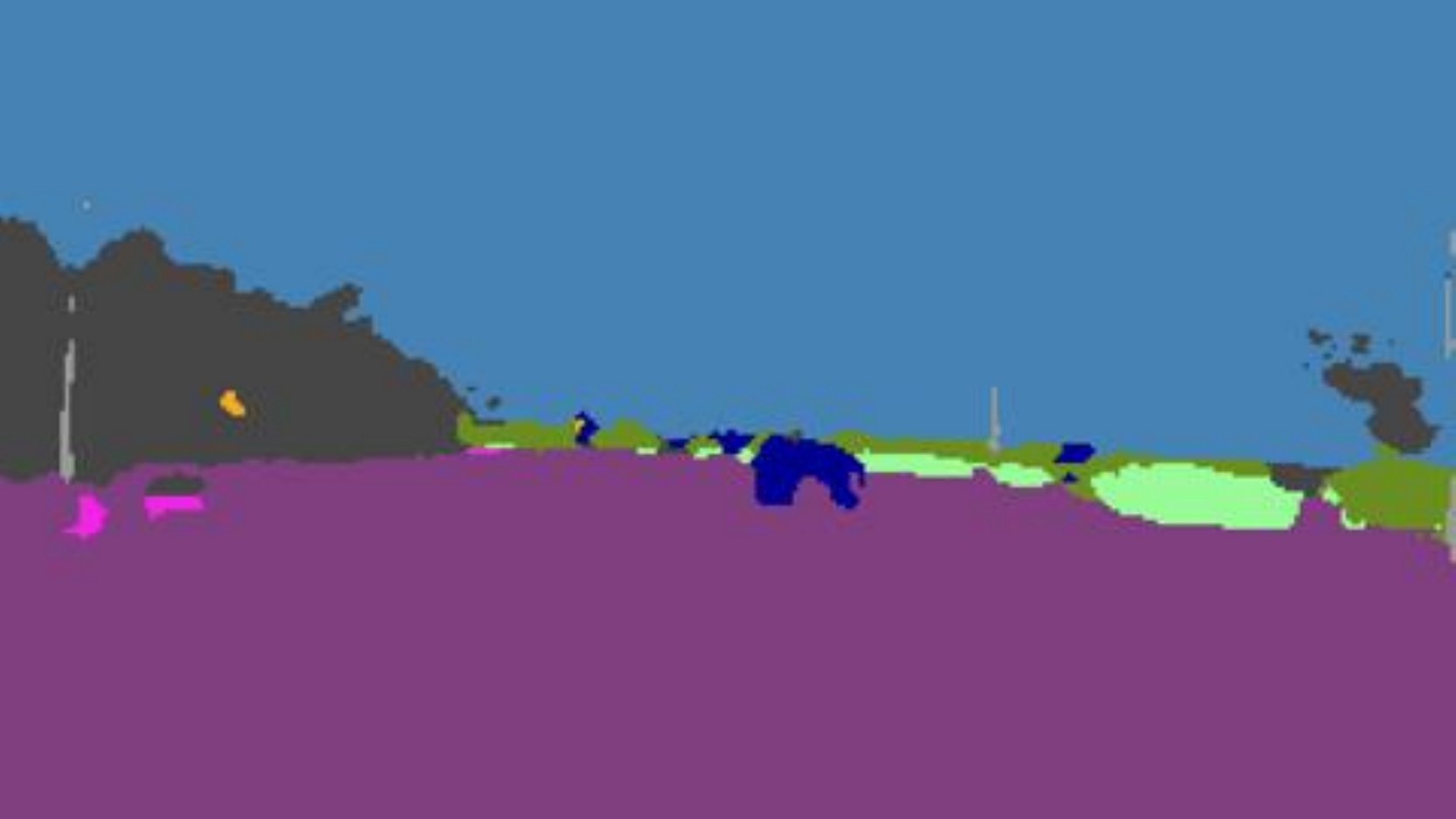} \\
    
    \includegraphics[width=\scaleSGOne]{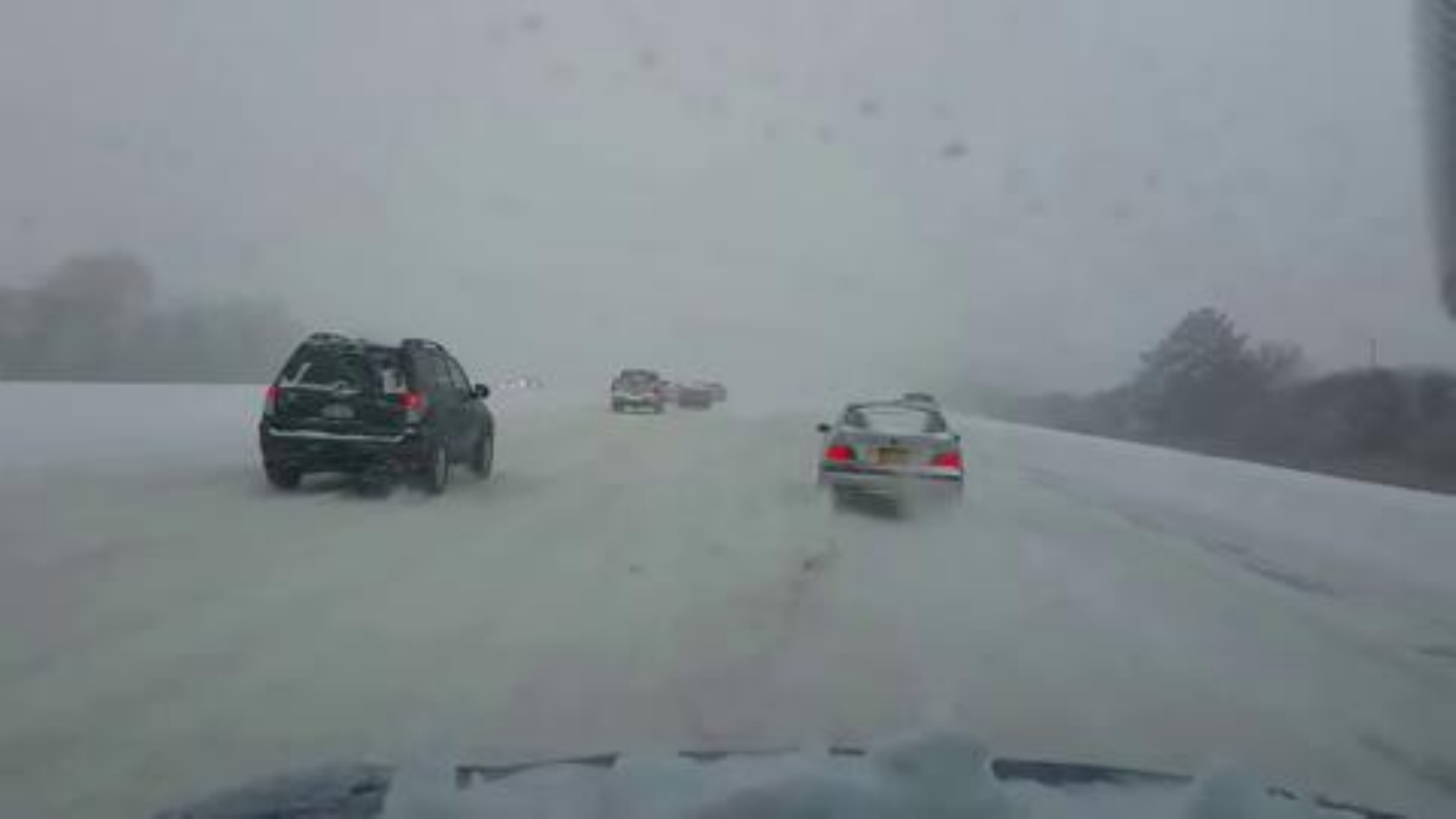} &
    \includegraphics[width=\scaleSGOne]{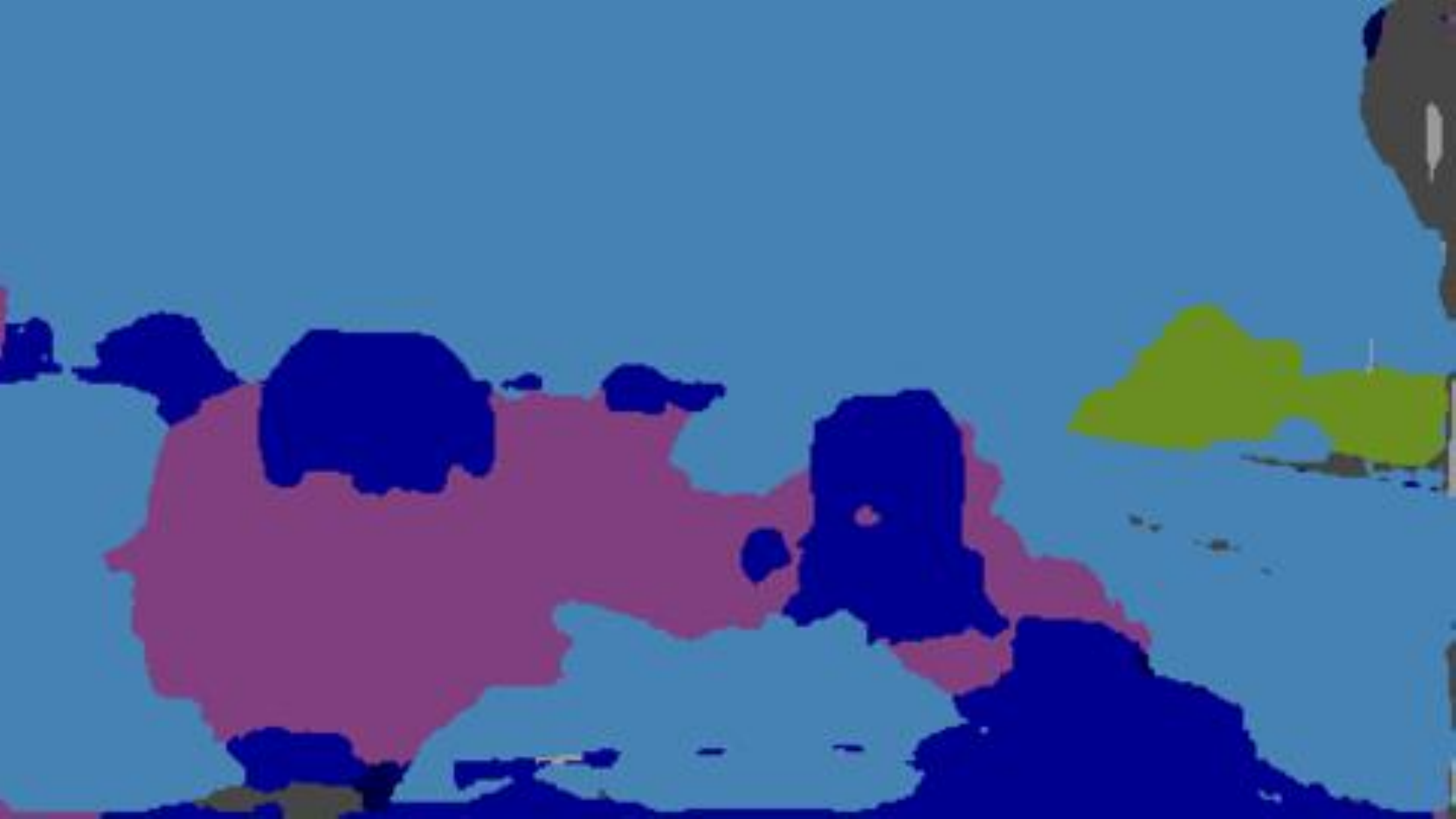} &
    \includegraphics[width=\scaleSGOne]{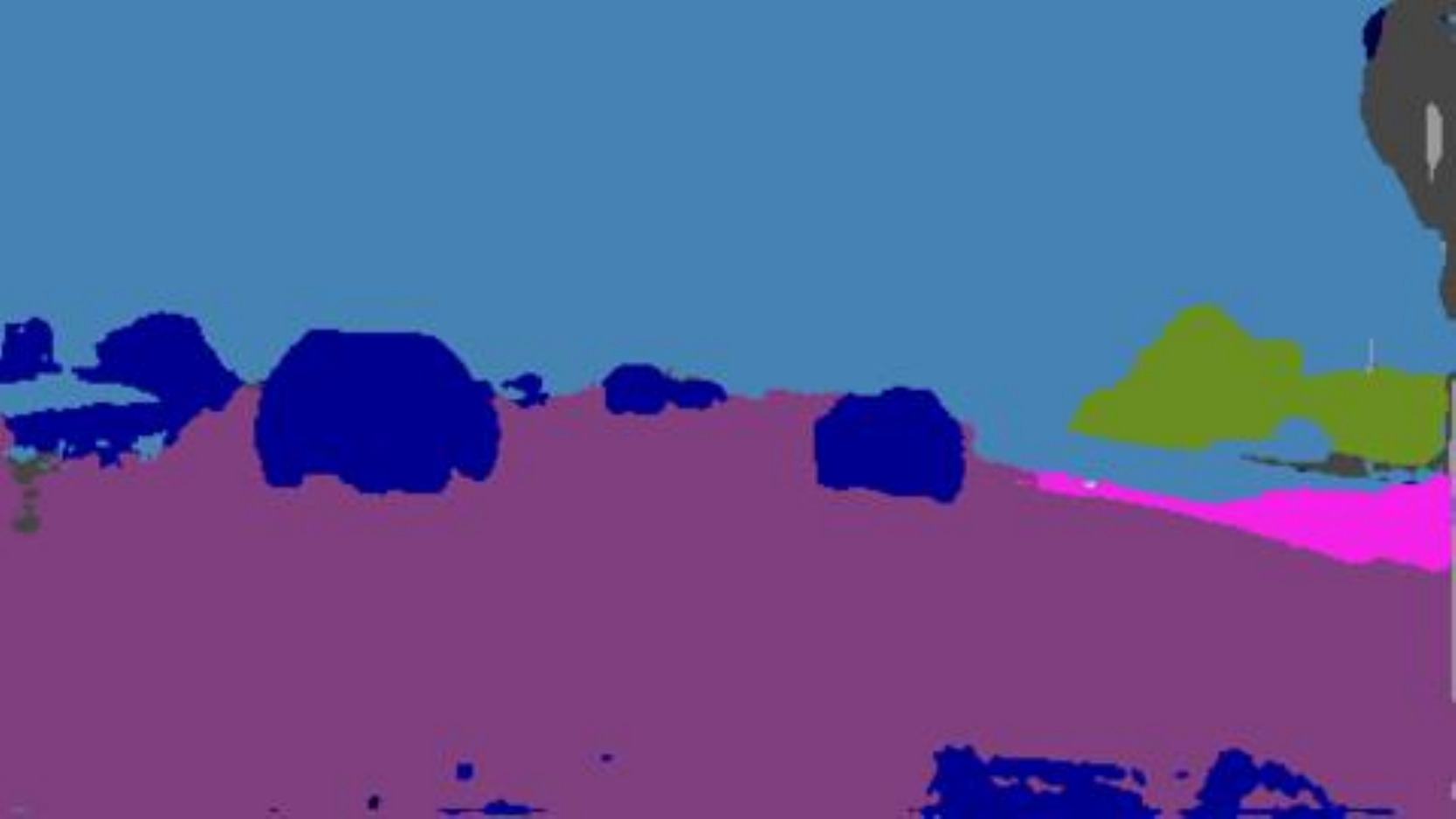} &
    \includegraphics[width=\scaleSGOne]{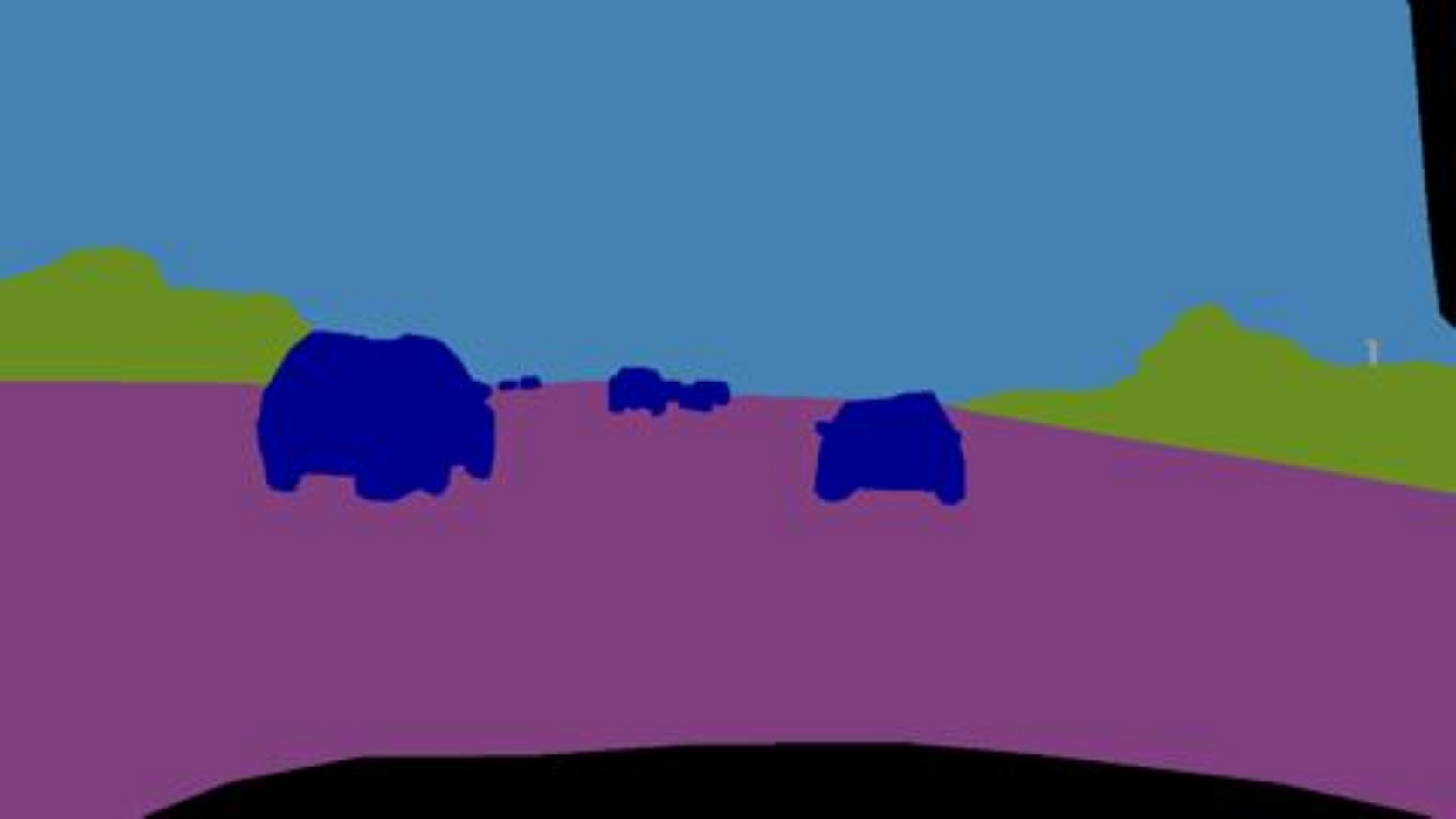} &
    \includegraphics[width=\scaleSGOne]{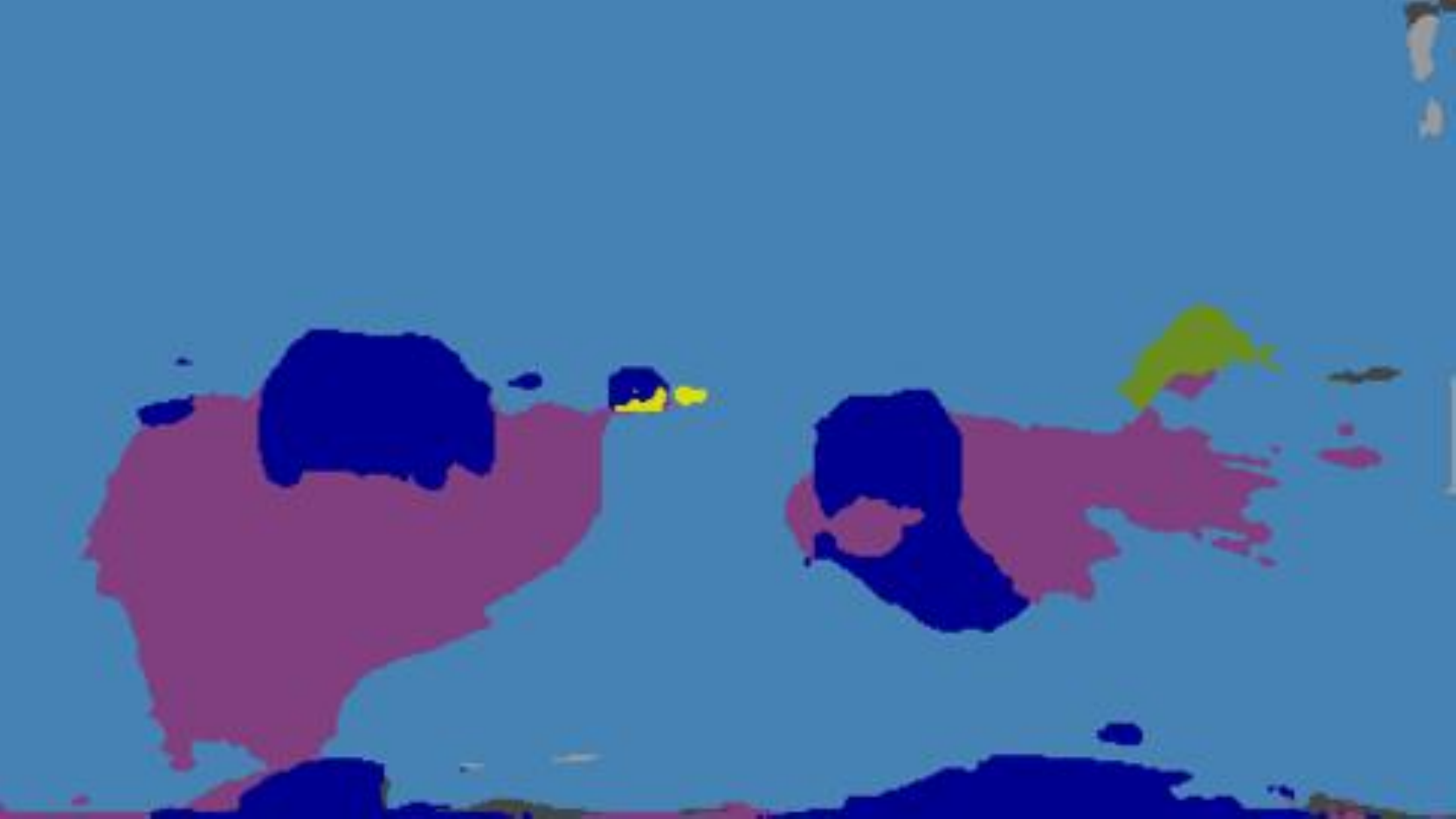} &
    \includegraphics[width=\scaleSGOne]{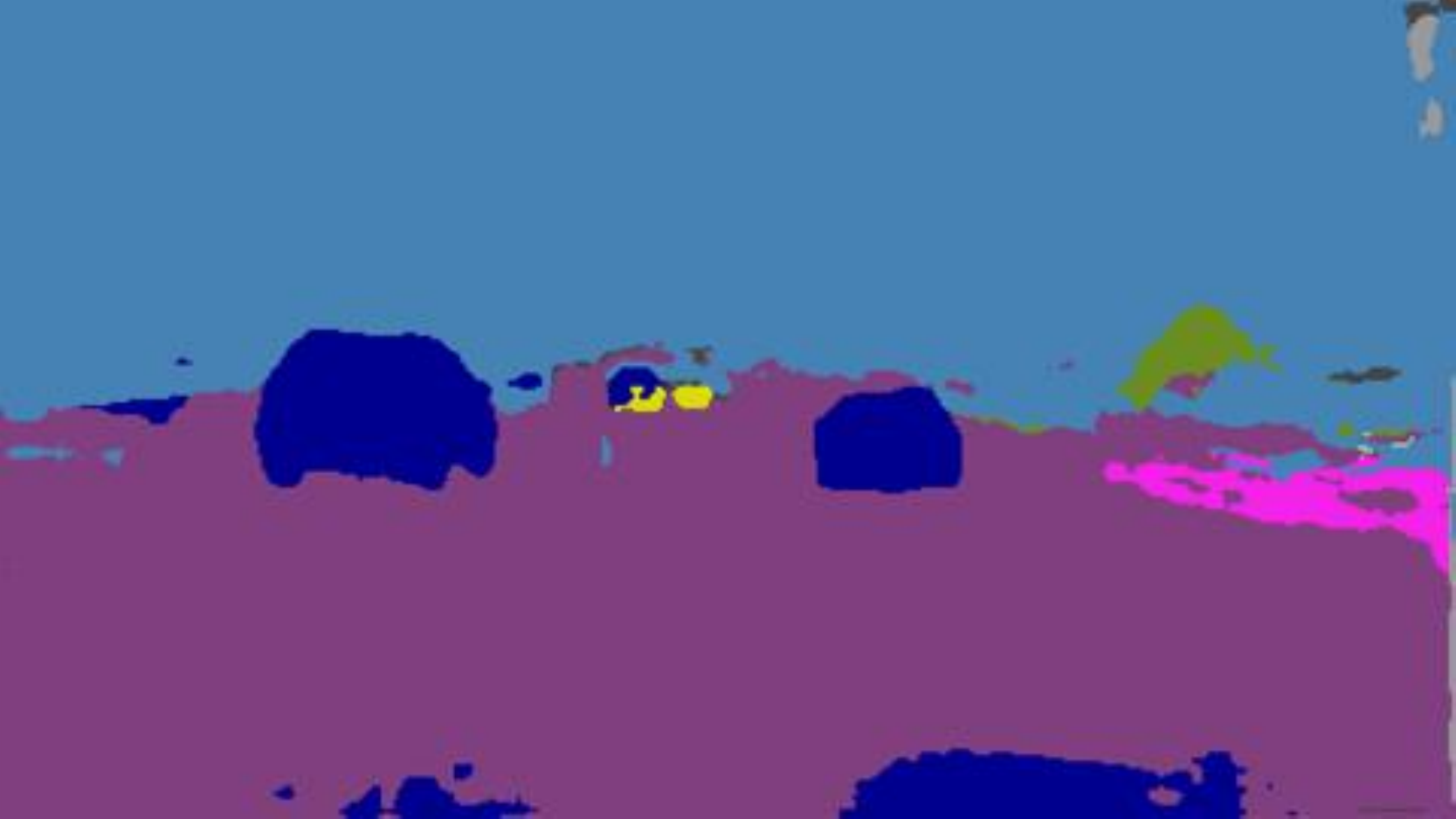} \\
    
    \includegraphics[width=\scaleSGOne]{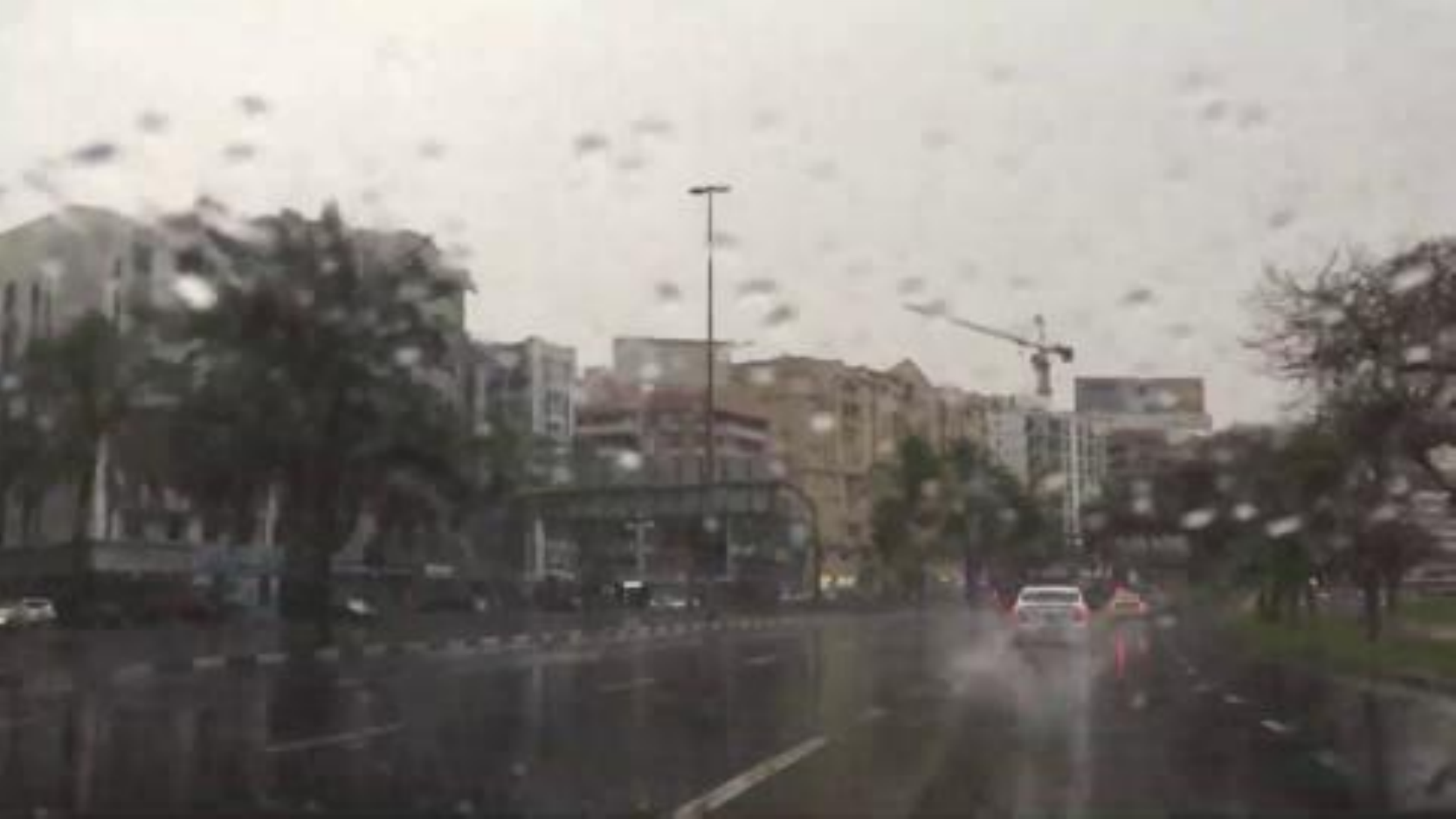} &
    \includegraphics[width=\scaleSGOne]{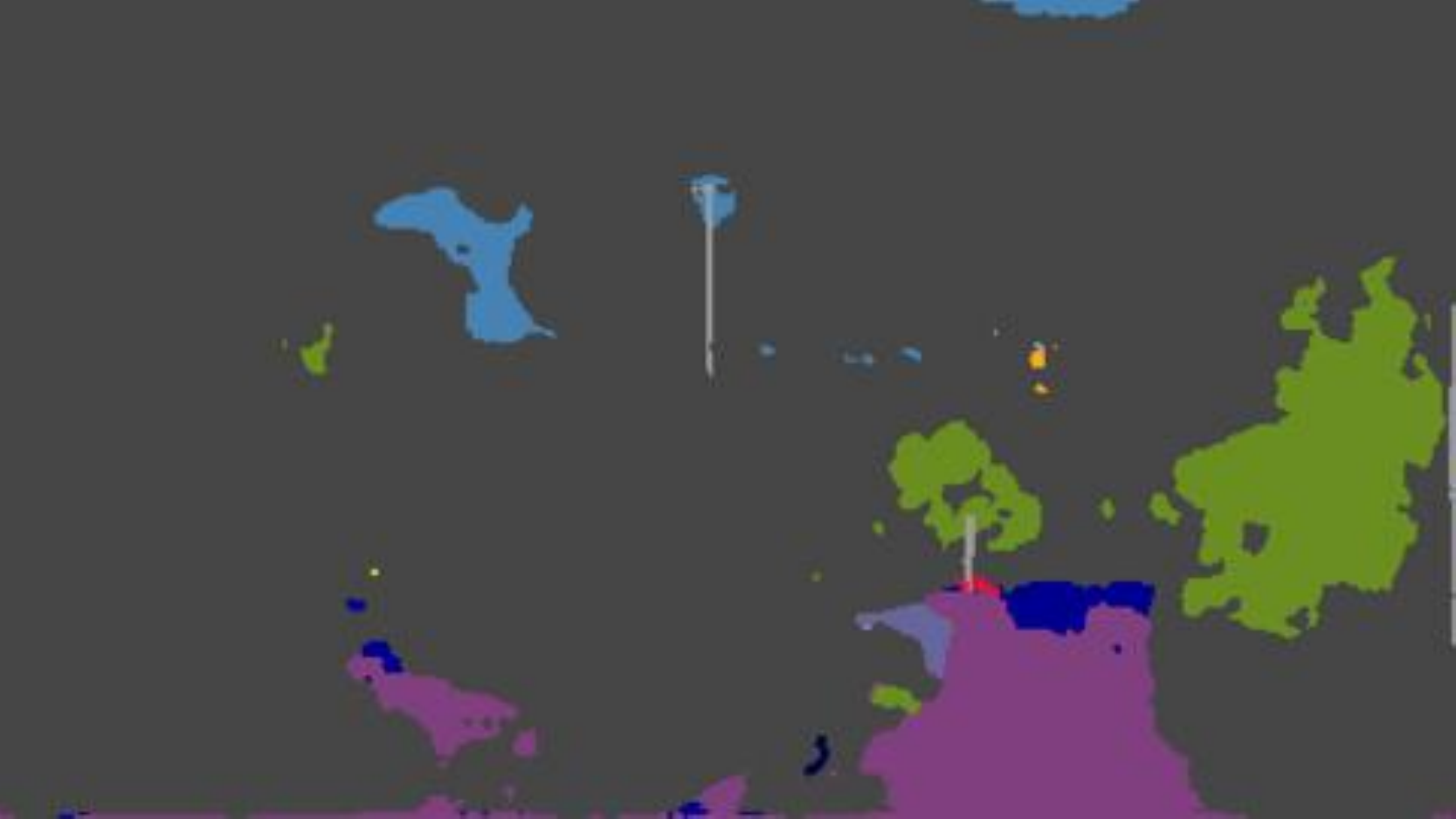} &
    \includegraphics[width=\scaleSGOne]{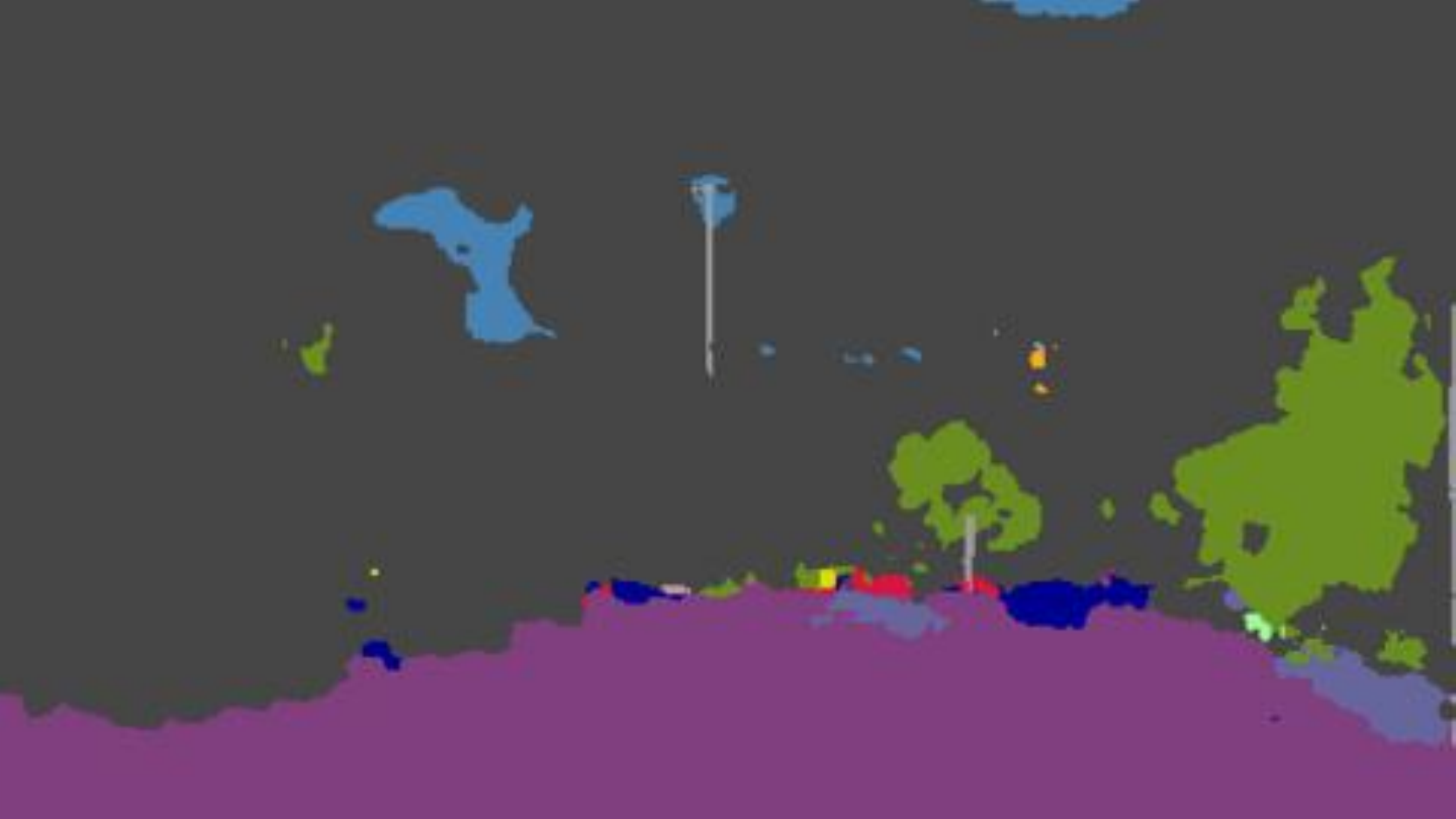} &
    \includegraphics[width=\scaleSGOne]{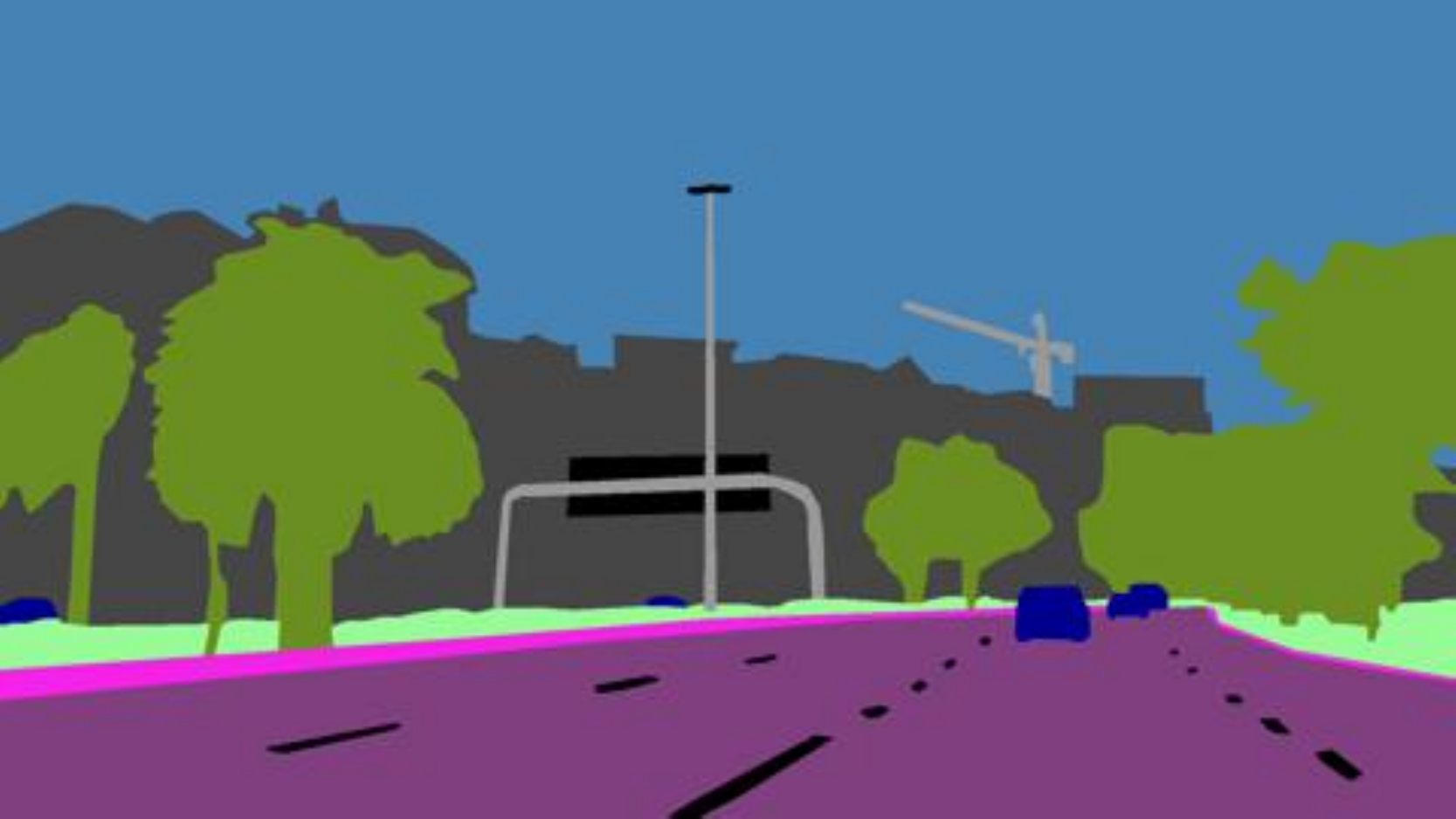} &
    \includegraphics[width=\scaleSGOne]{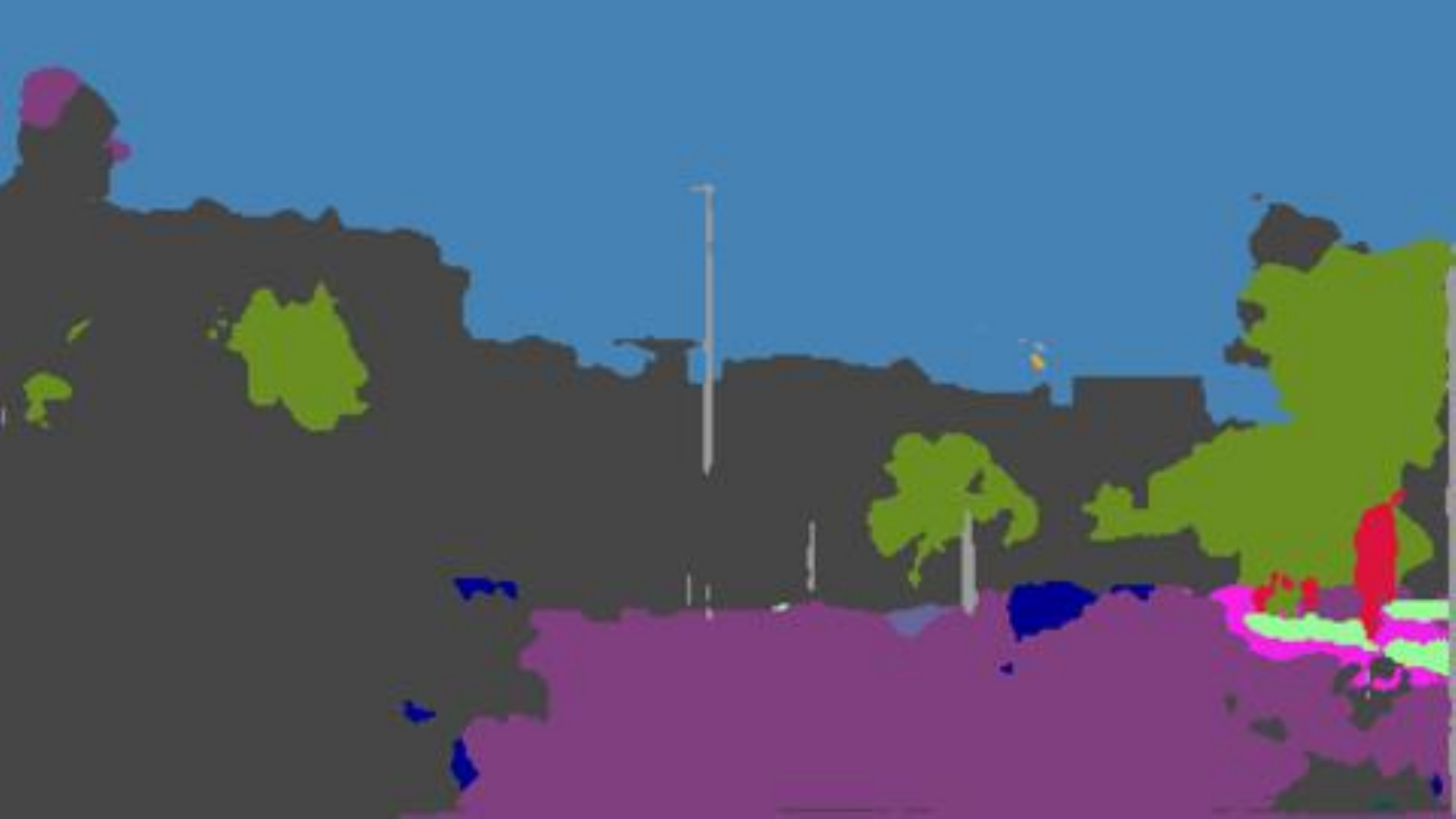} &
    \includegraphics[width=\scaleSGOne]{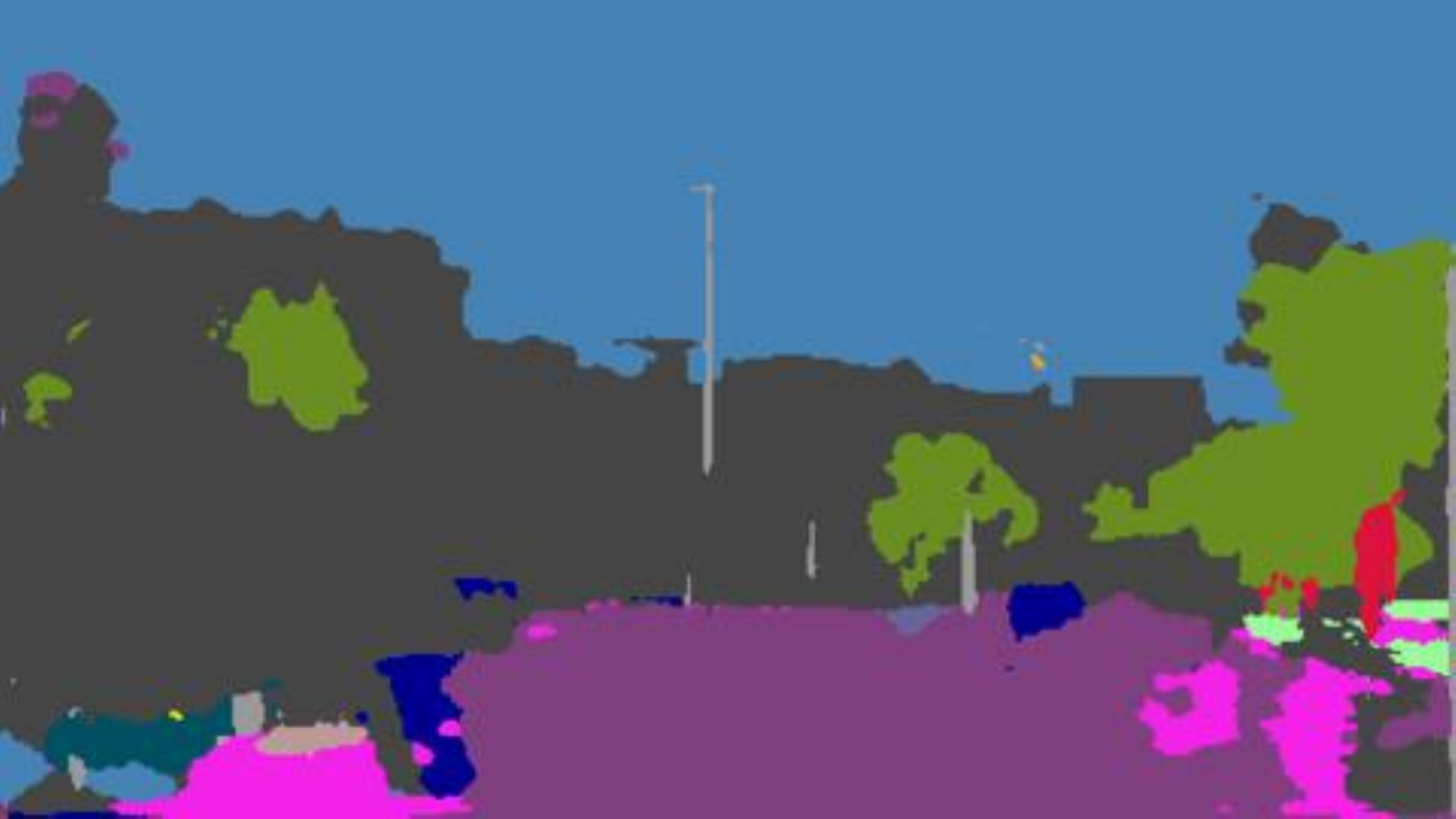} \\
    
    \includegraphics[width=\scaleSGOne]{images/Qualitative/Road_Segmentation/Good_Results/BDD_Night/c776d10b-06d930cf/c776d10b-06d930cf.pdf} &
    \includegraphics[width=\scaleSGOne]{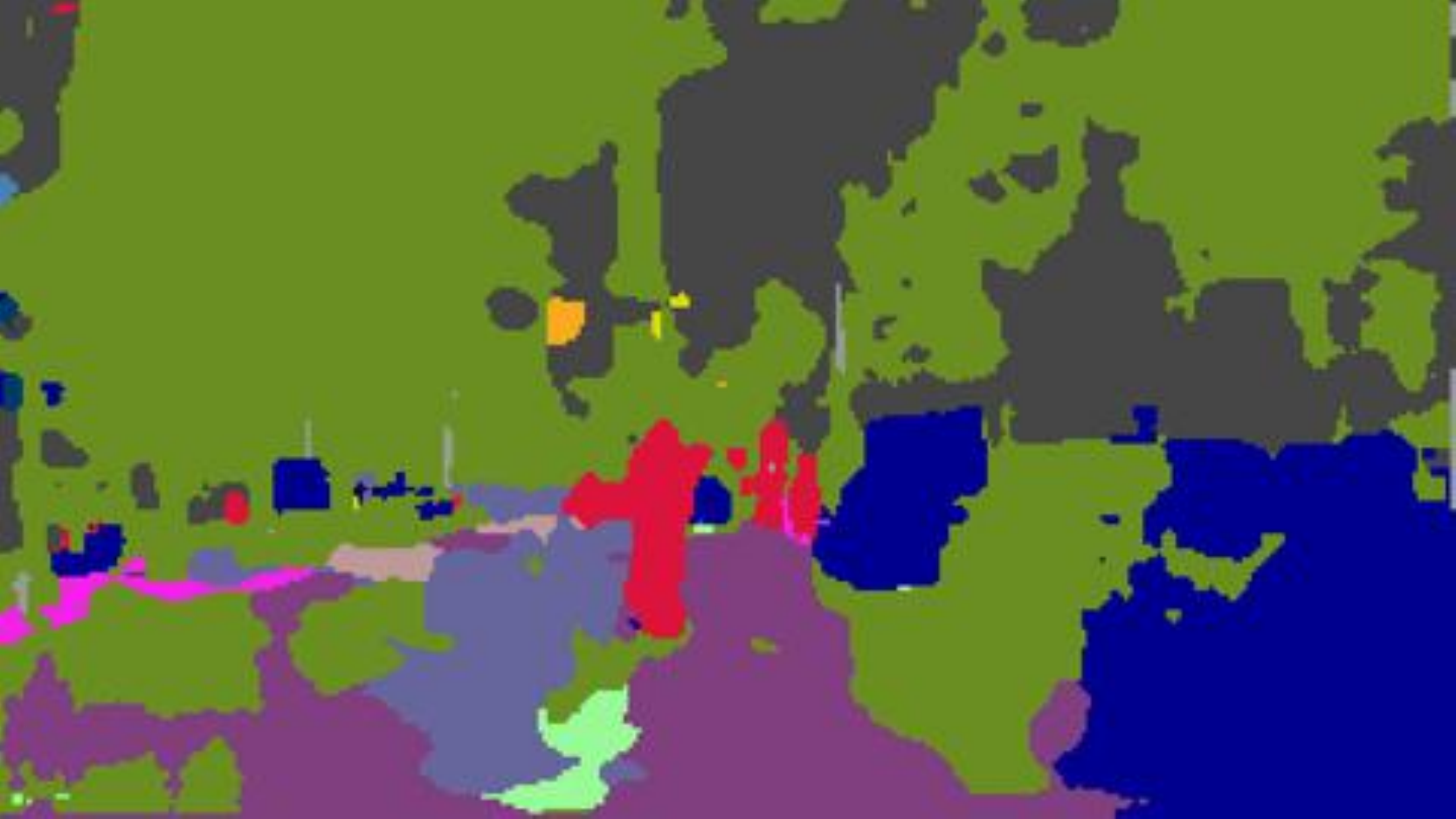} &
    \includegraphics[width=\scaleSGOne]{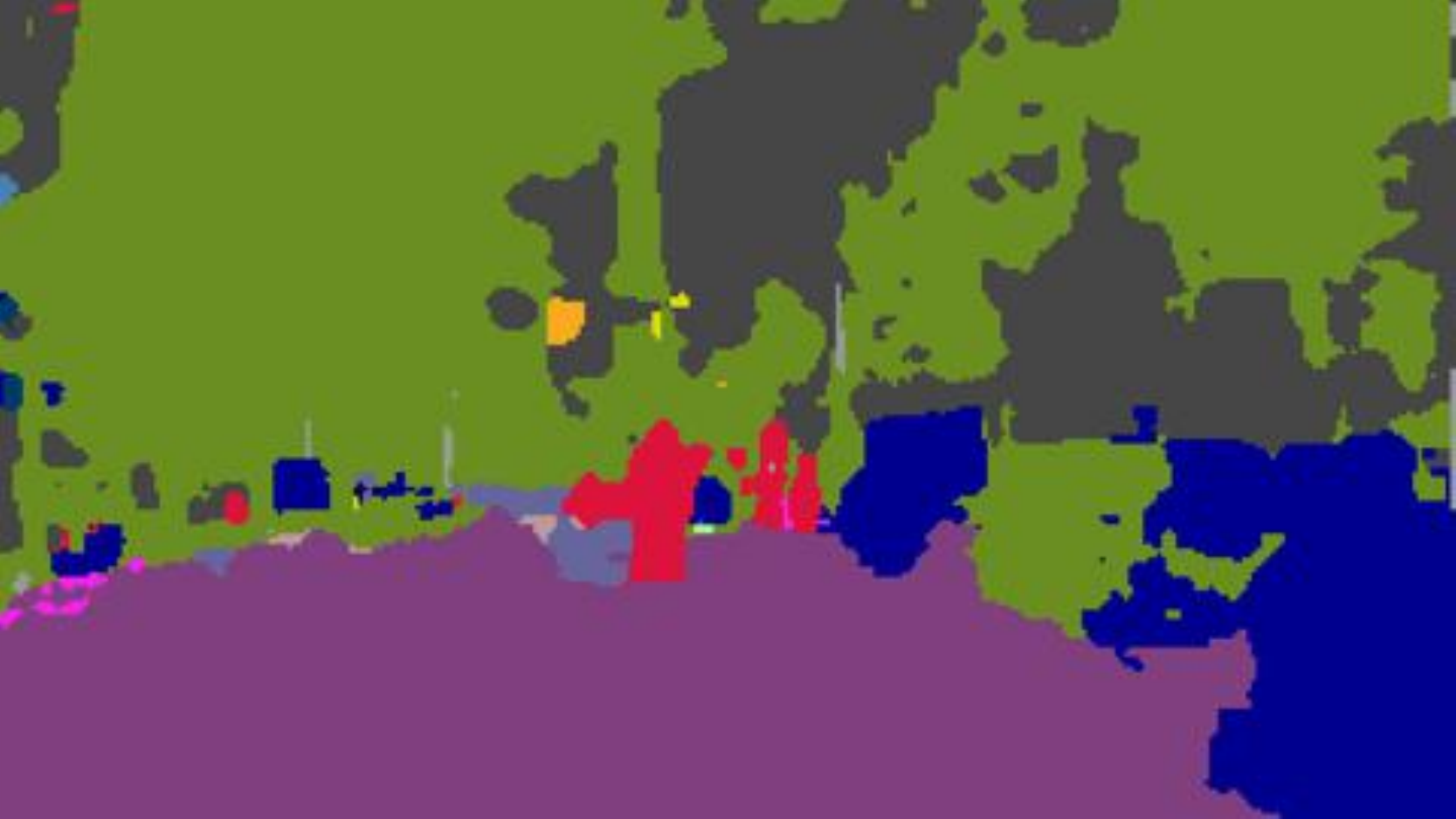} &
    \includegraphics[width=\scaleSGOne]{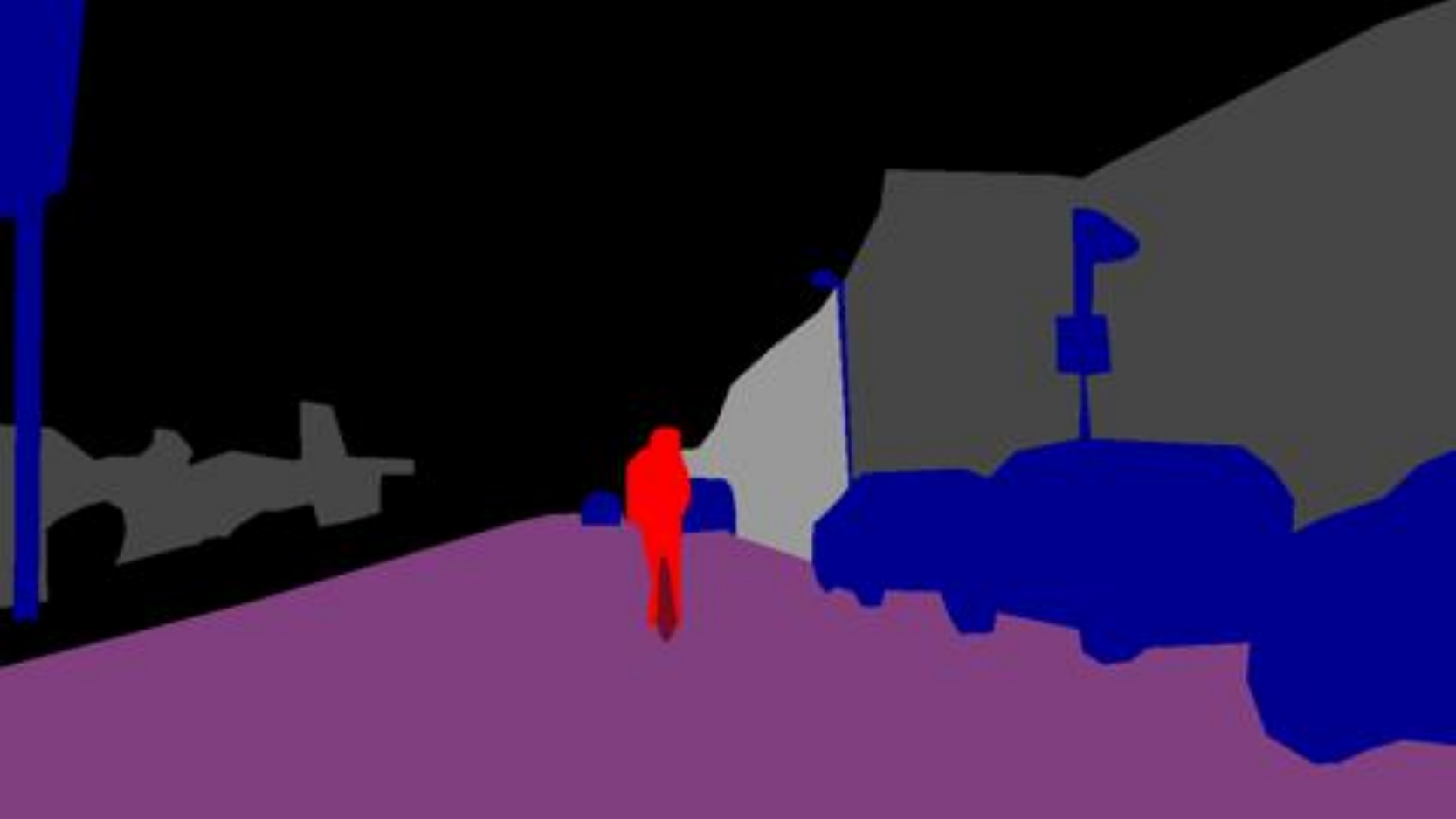} &
    \includegraphics[width=\scaleSGOne]{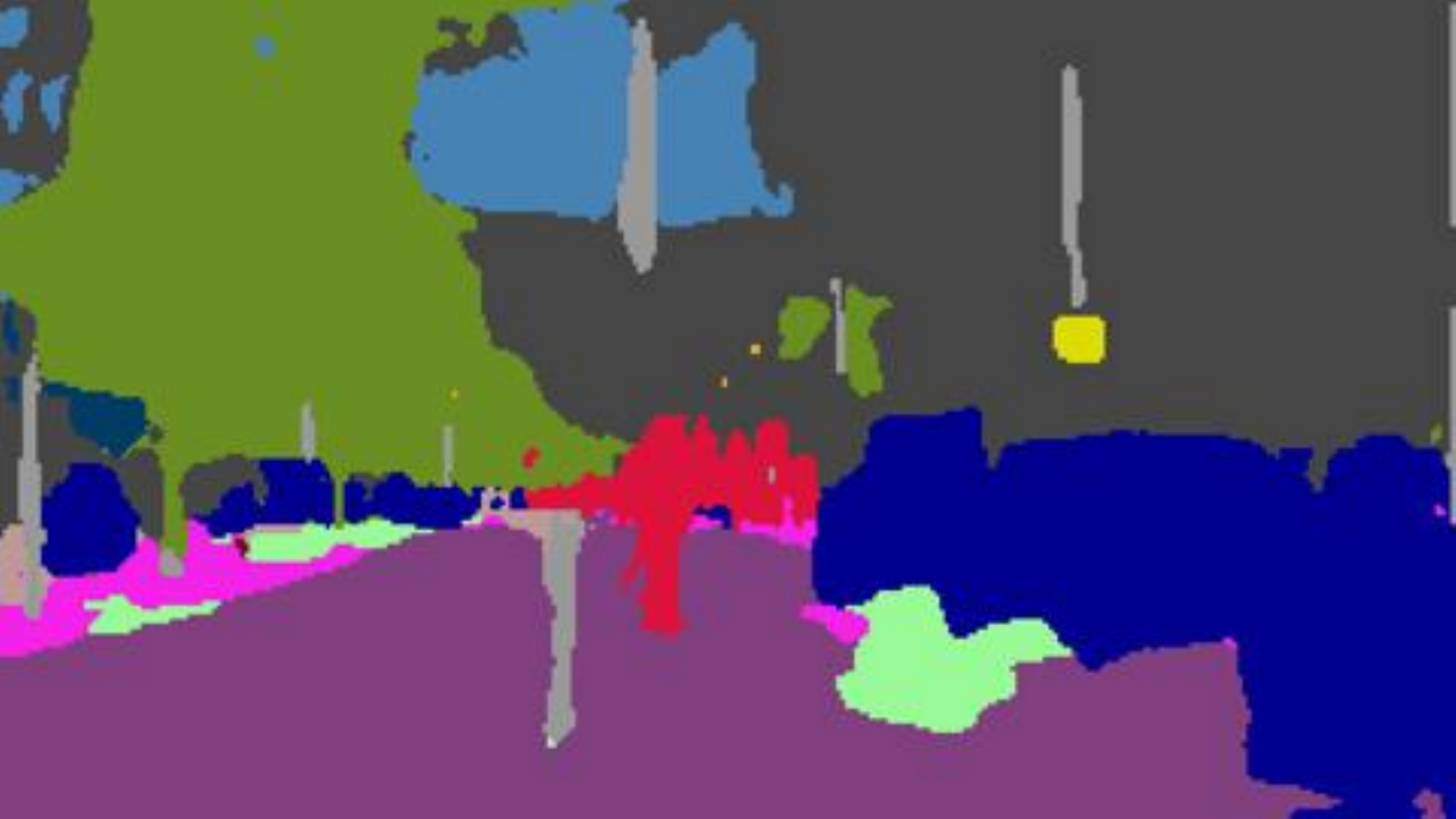} &
    \includegraphics[width=\scaleSGOne]{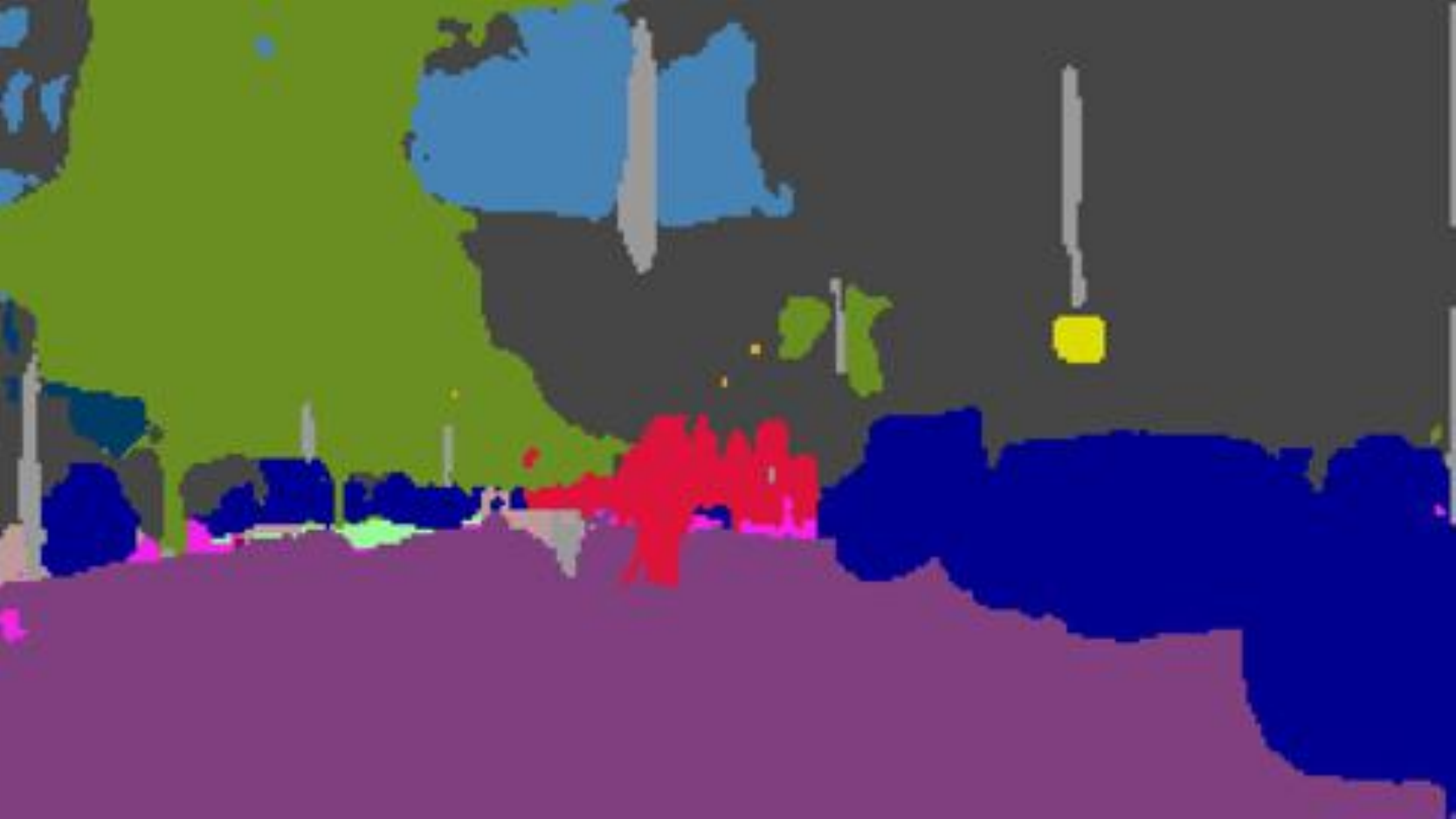} \\
    \midrule
    
    \includegraphics[width=\scaleSGOne]{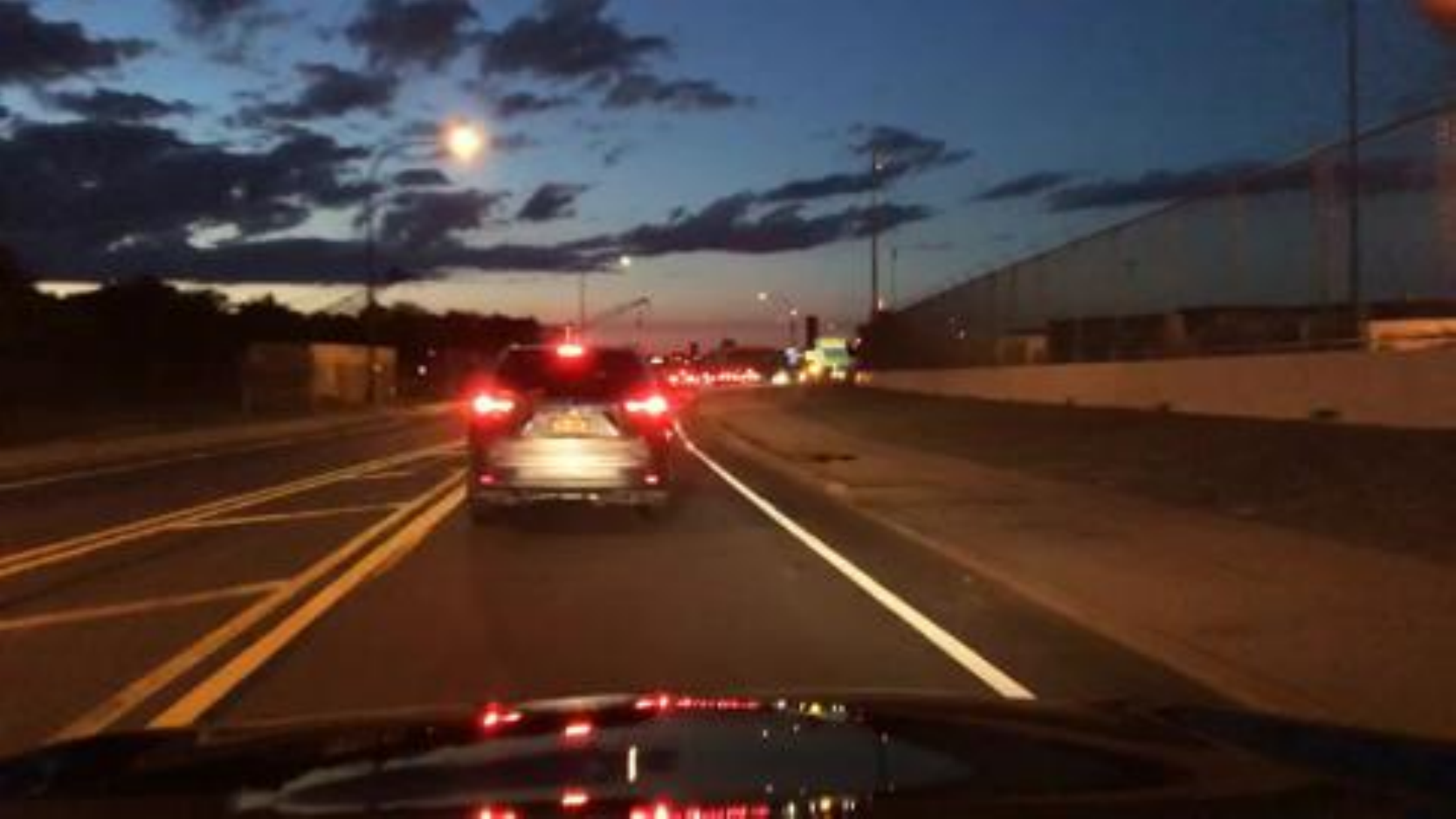} &
    \includegraphics[width=\scaleSGOne]{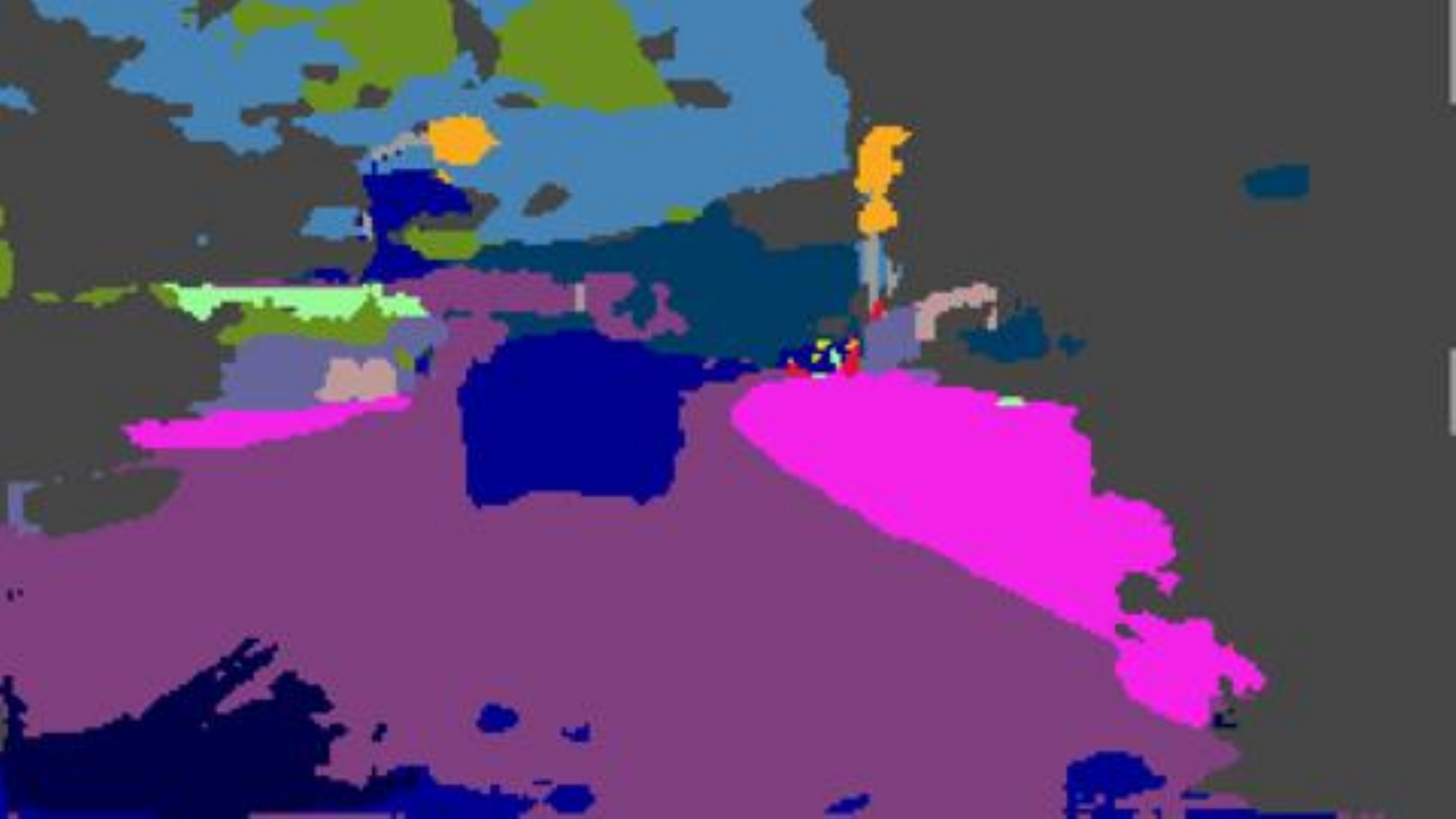} &
    \includegraphics[width=\scaleSGOne]{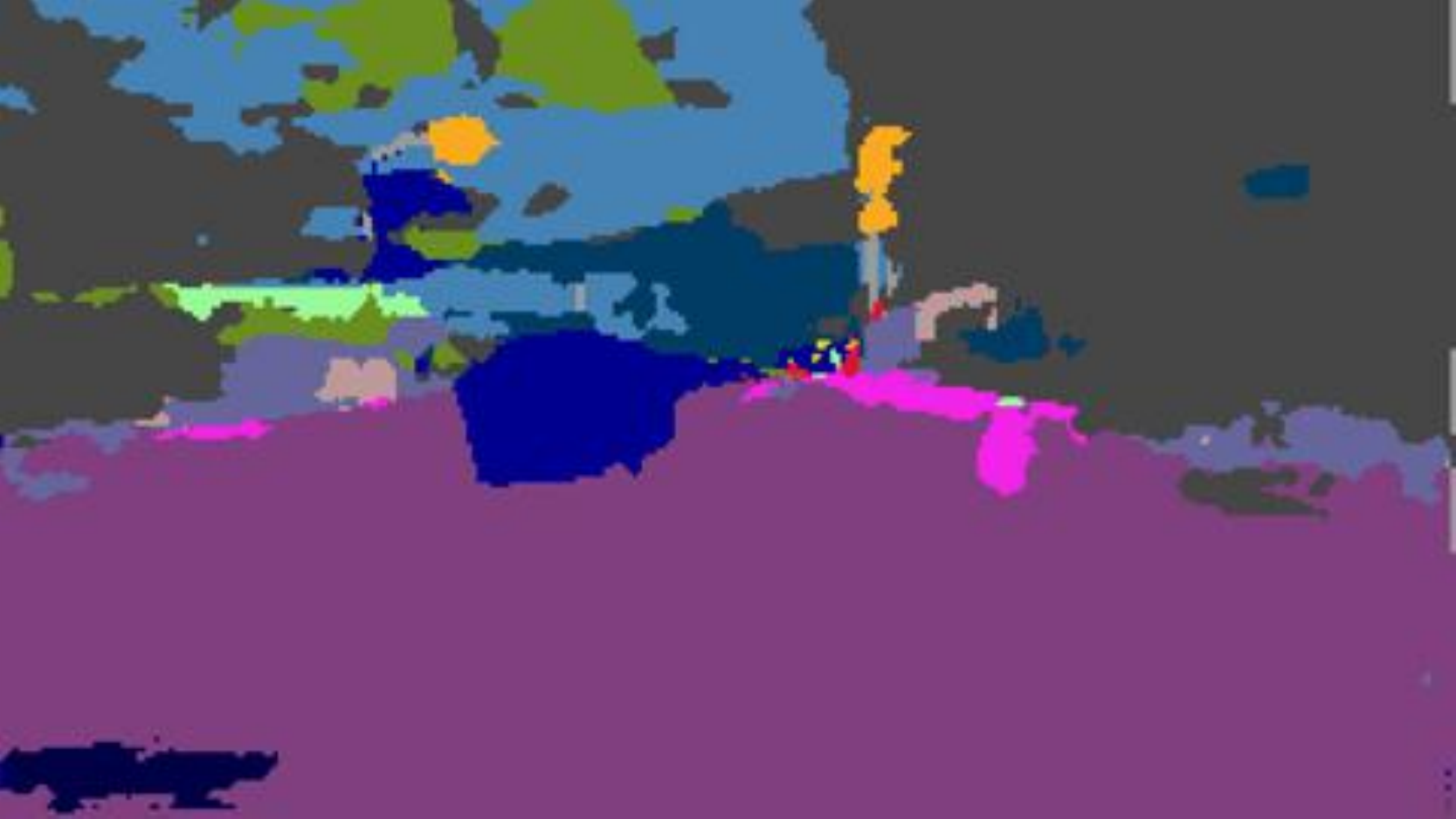} &
    \includegraphics[width=\scaleSGOne]{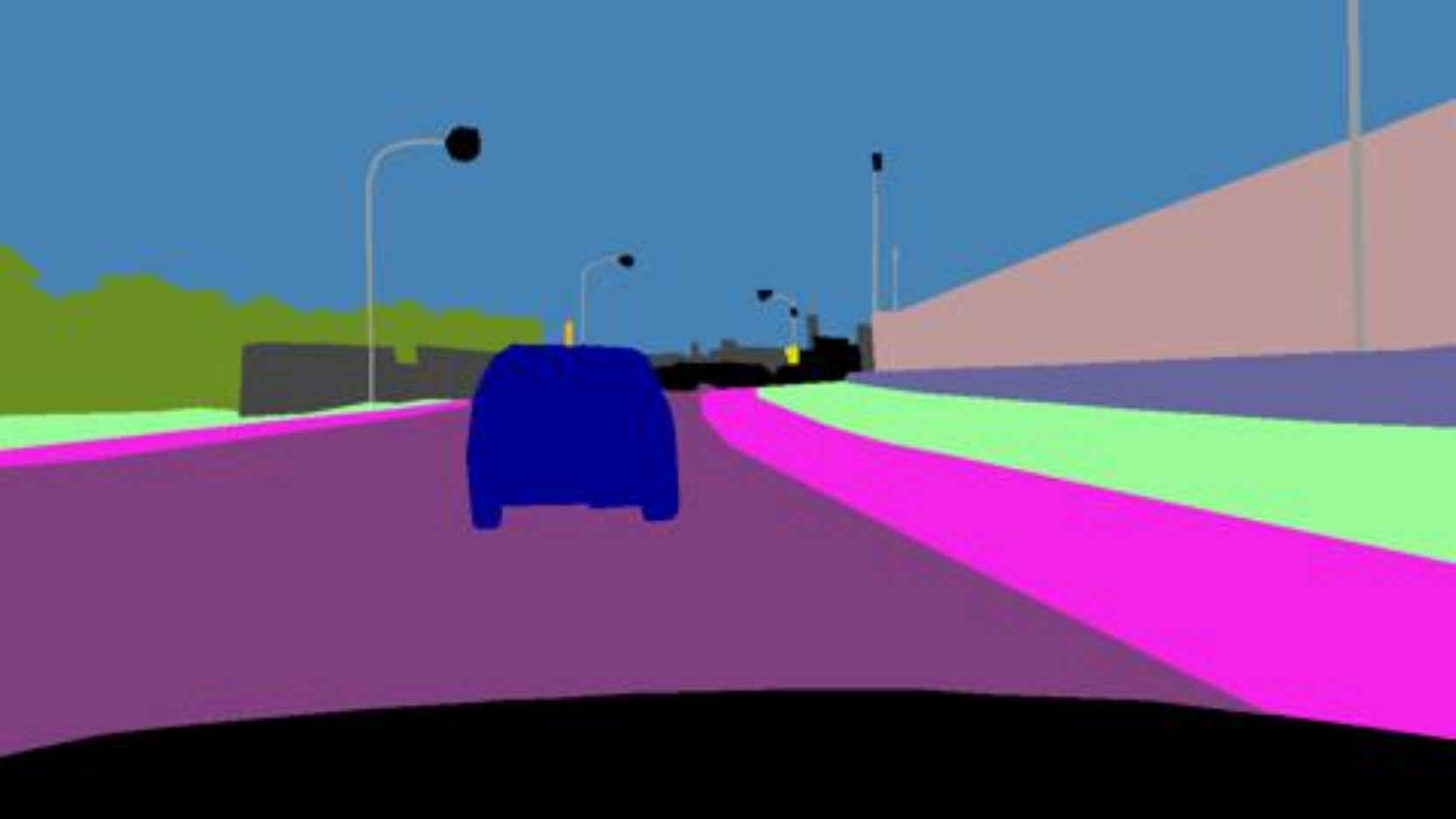} &
    \includegraphics[width=\scaleSGOne]{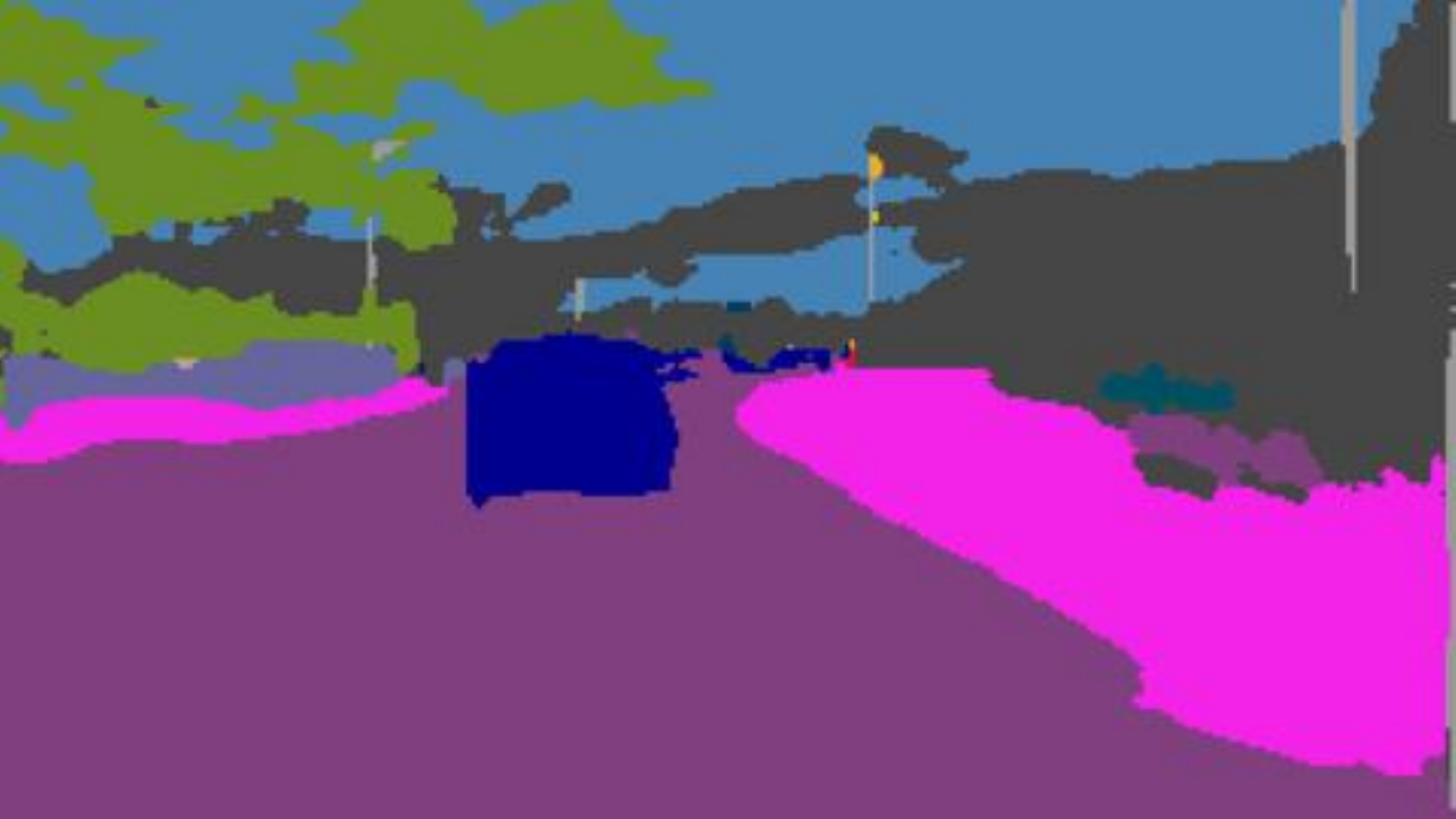} &
    \includegraphics[width=\scaleSGOne]{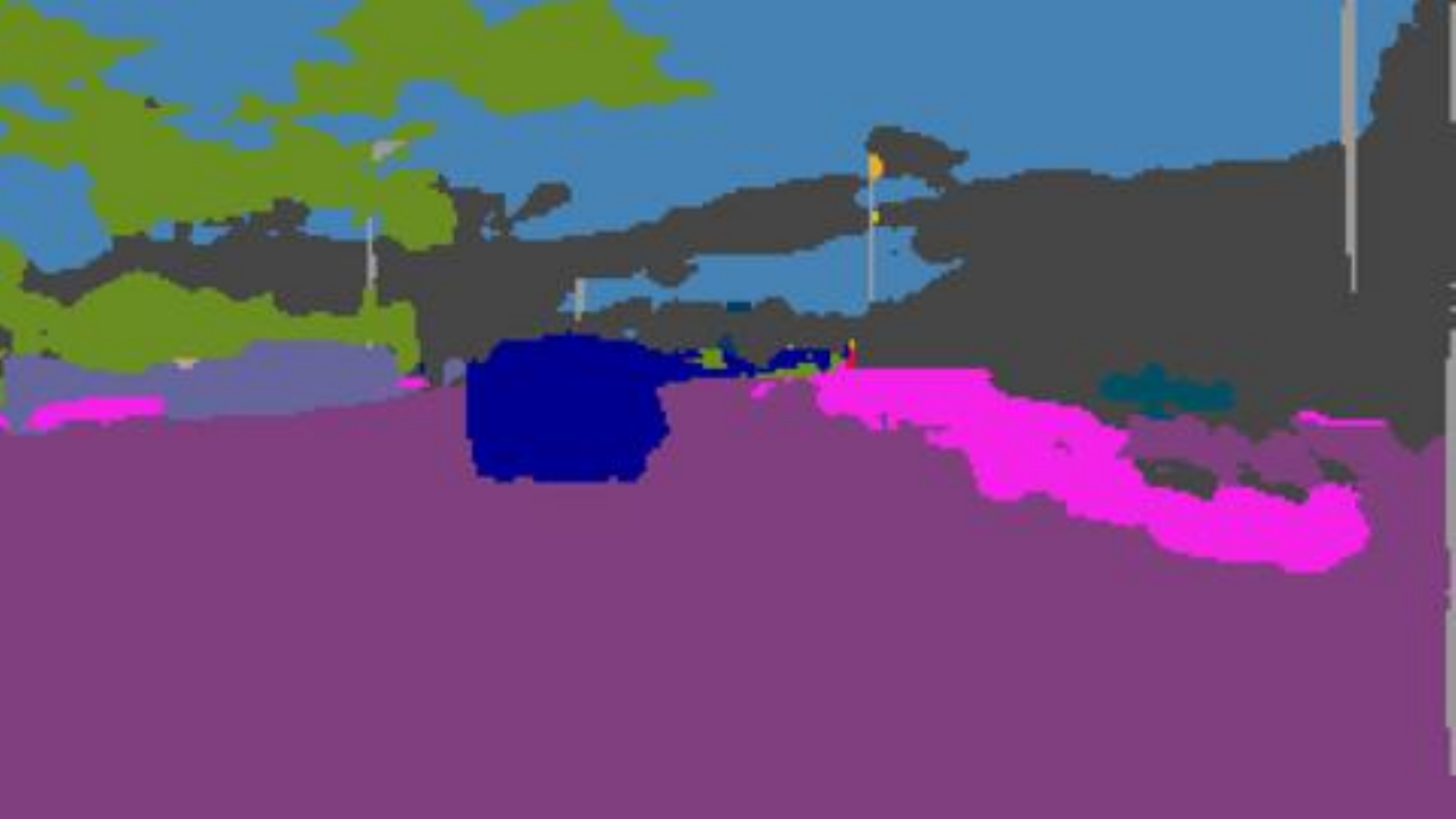} \\

    \multicolumn{6}{c}{\includegraphics[scale=0.24]{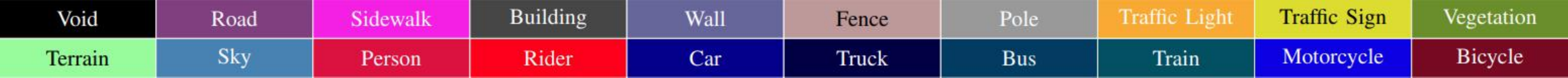}} 
    \end{tabular}
    \caption{Qualitative segmentation results for RefineNet based networks. In general our approach leads to improved road segmentation in various challenging domains - even providing further improvement to current SOTA domain adaptation approaches. A failure case is presented in row 6.}
    \label{fig:QualResDL}
\end{figure*}

\subsection{Prior-Based Road Segmentation}
Table~\ref{tab:singleTable} demonstrates that our proposed system is able to improve the performance of all tested networks across multiple unseen challenging domains. Notably, we show that our system is also able to complement and improve the performance of current state-of-the-art domain adaptation methods which already yield considerable performance gains compared to using the pre-trained segmentation networks on the raw query images. Column 3 and 5 of Figure~\ref{fig:QualResDL} show that we are able to improve road segmentation similarly to MGCDA without the requirement of labelled training data.
Performance improvements are achieved for all techniques, but are particularly significant where the baseline open-loop road segmentation is poor, such as for DeepLabV3+ on the BDD100K Night dataset. Figure~\ref{fig:QualResDL} qualitatively shows the explicit contribution of the various system components. In particular it shows that domain adaptation from MGCDA even increases robustness in unseen domains, but our method complements this to provide further improvements.

We include comparisons with a simple baseline system, where the average road position over the reference set is `copied and pasted' to the query, and with an alternative version of our system but with a naive rather than Bayesian fusion component. The similar place template approach provides a more accurate road prior than the simple baseline, and our full system typically outperforms both of these overall. Interestingly, the approach of using only the similar place prior outperforms our Bayesian fusion method for the Wild Dash night, rain and BDD night datasets using the DeepLab network. This is likely because the baseline performance is so poor, so any other similar place in this case has a better segmentation result than the open loop performance on the query image. Here, the Bayesian update combines the the similar place prior's prediction and the very poor open loop query segmentation, which deteriorates performance.

In the following subsections, we discuss the performance benefit of averaging the top 5 similar place candidates and some of the frame by frame results. 

\begin{figure}
    \centering
    \begin{tabular}{cc}
    \includegraphics[width=0.22\textwidth]{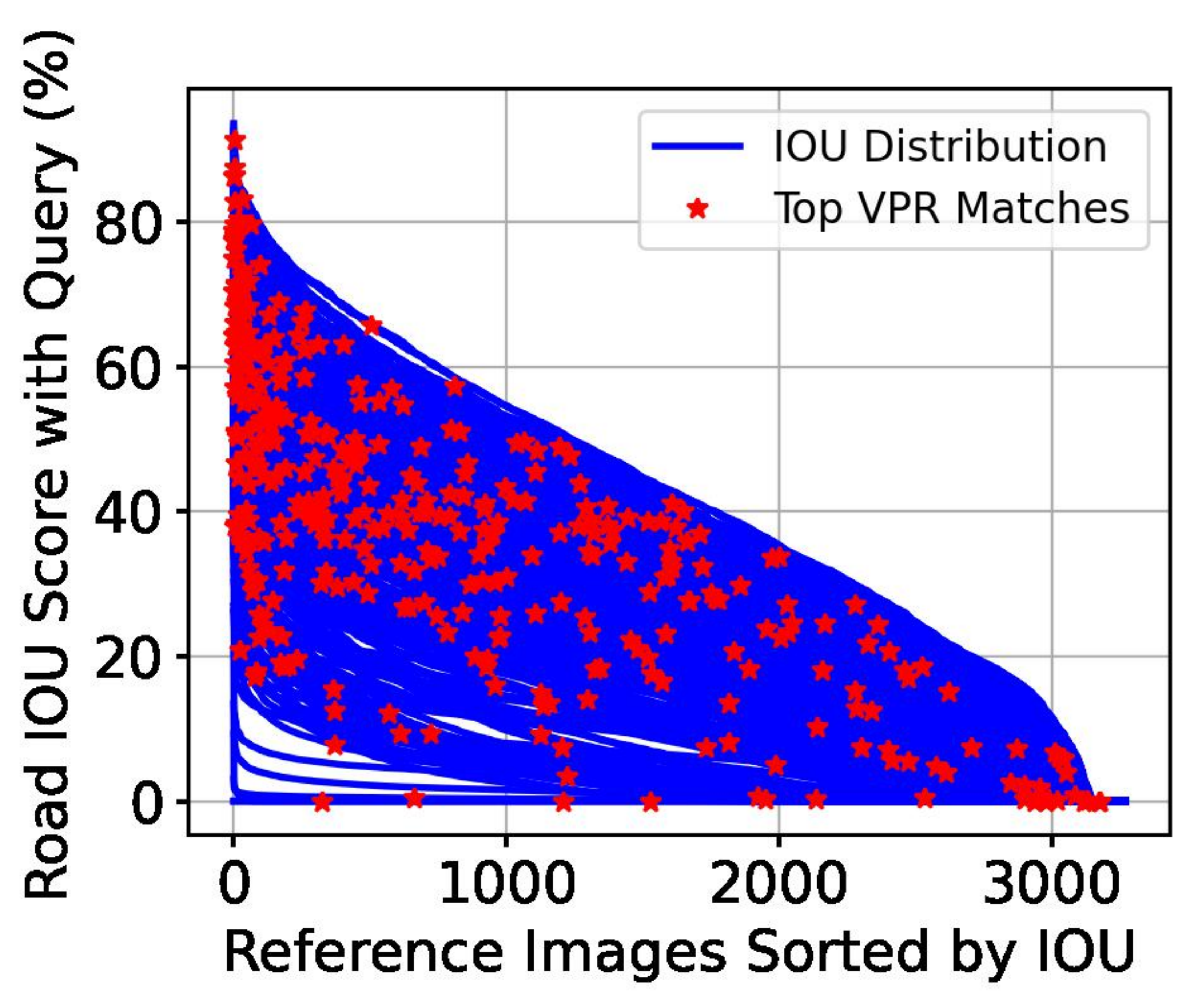} &
    \includegraphics[width=0.25\textwidth]{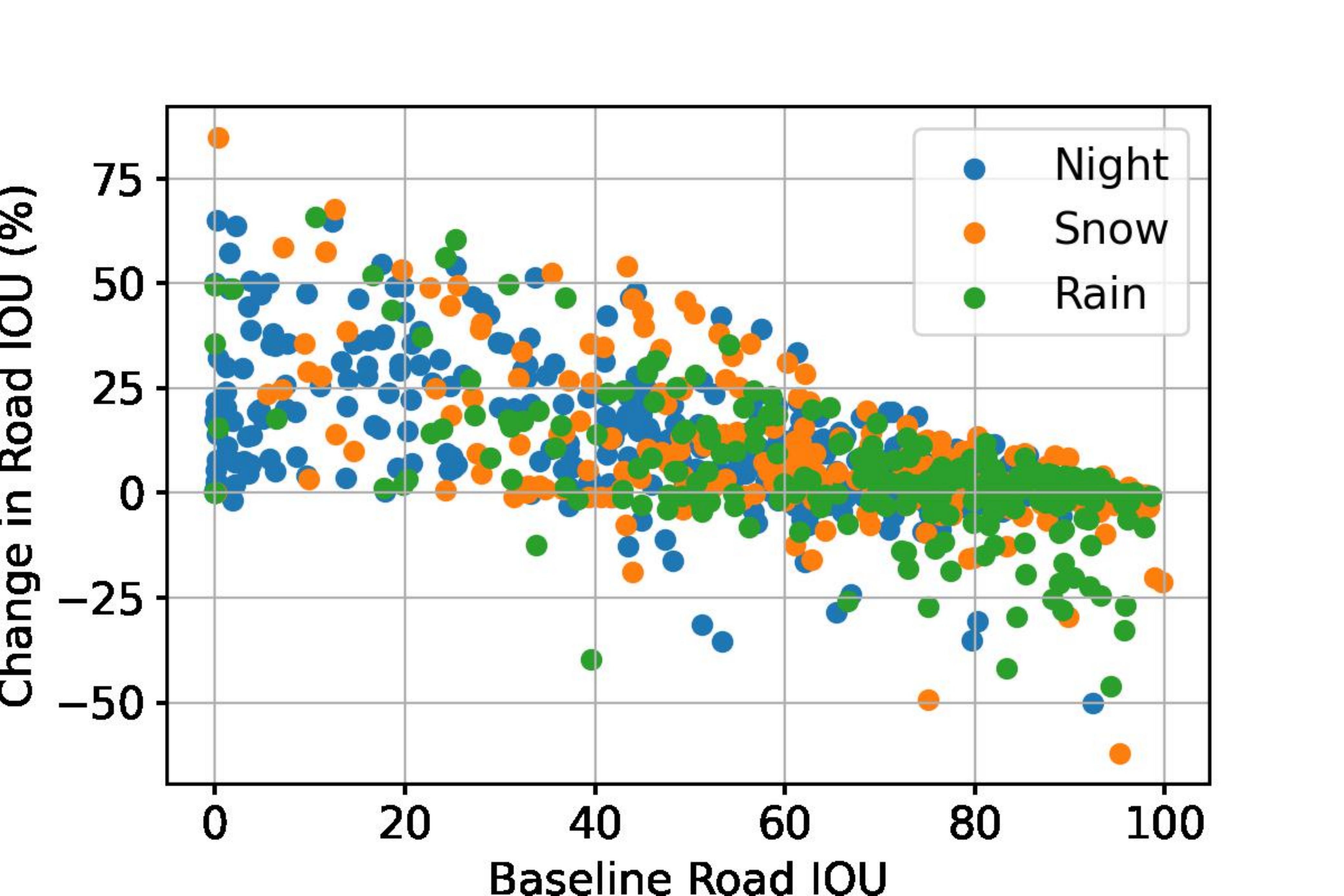} \\
    (a) & (b) \\
    \end{tabular}
    \caption{(a) VPR similarity performance across the WildDash 2 Night, showing that top-ranked VPR matches tend to be placed with a high Road IOU score with the query image. (b) Per frame results for the RefineNet baseline Road IOU plotted against the change in IOU using our method, indicating our method has a particularly significant effect on low baselines.}
    \label{fig:VPRSimilarity+BaseVsImp}
\end{figure}

\paragraph*{Ablating the Number of `$k$' Retrieved Places}
During early development we observed that the top-ranked place match chosen by HybridNet is not always the \textit{most} structurally similar image (Figure \ref{fig:VPRSimilarity+BaseVsImp}(a))  from amongst the top-ranked matching candidates. To investigate the effect of averaging over the top \textit{k} matches, we used ground truth annotations to compute the IOU scores between each query and all references images, and visualized the similarity distribution (Figure~\ref{fig:VPRSimilarity+BaseVsImp}(a), noting some data points show a Road IOU of 0 because some WildDash imagery is off-road with no road labelled). While the top VPR matches are heavily weighted towards the left side of the graph, they are not always the absolute highest rank - explaining why generating a prior ($x_{n}^s$) from averaging the top 5 similar places provides an advantage over simply using the top match. 

\paragraph*{Frame by Frame Performance}
Figure \ref{fig:VPRSimilarity+BaseVsImp}(b) illustrates the change in road IOU against base segmentation IOU across the WildDash 2 datasets after applying our method to RefineNet predictions. Our method substantially improves cases where road segmentation fails catastrophically (e.g. top-left hand side) while not compromising performance when RefineNet yields acceptable segmentation performance (right hand side) overall. Failure cases still occur, typically caused by VPR not finding a sufficiently similar prior place from the reference set. Another failure case occurs when classes adjacent to the road, such as sidewalk, become eroded by the new road prediction, as in Figure~\ref{fig:QualResDL}, row 6.

\begin{figure}
    \centering
    \begin{tabular}{cc}
    \includegraphics[width=0.225\textwidth]{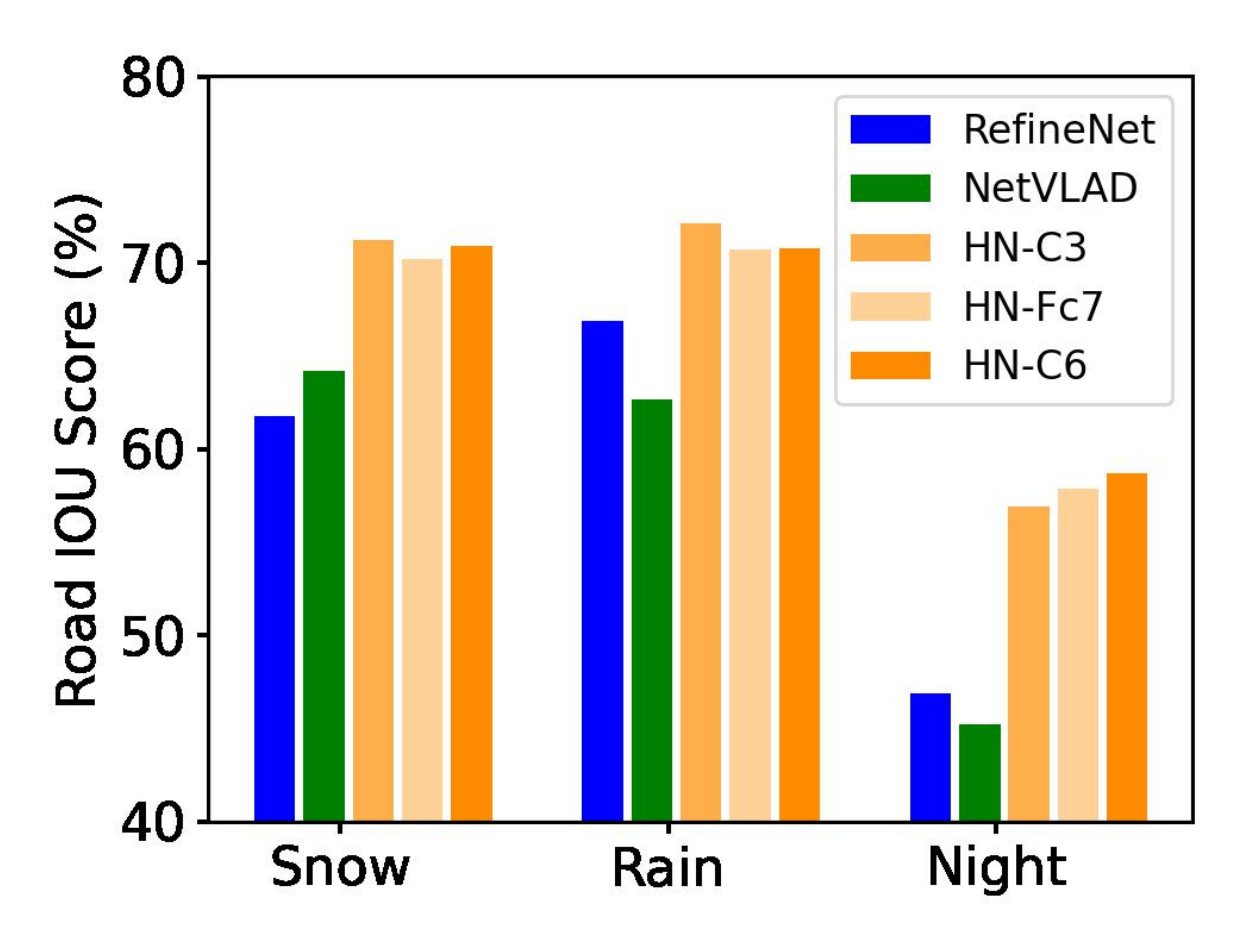}  &
    \includegraphics[width=0.225\textwidth]{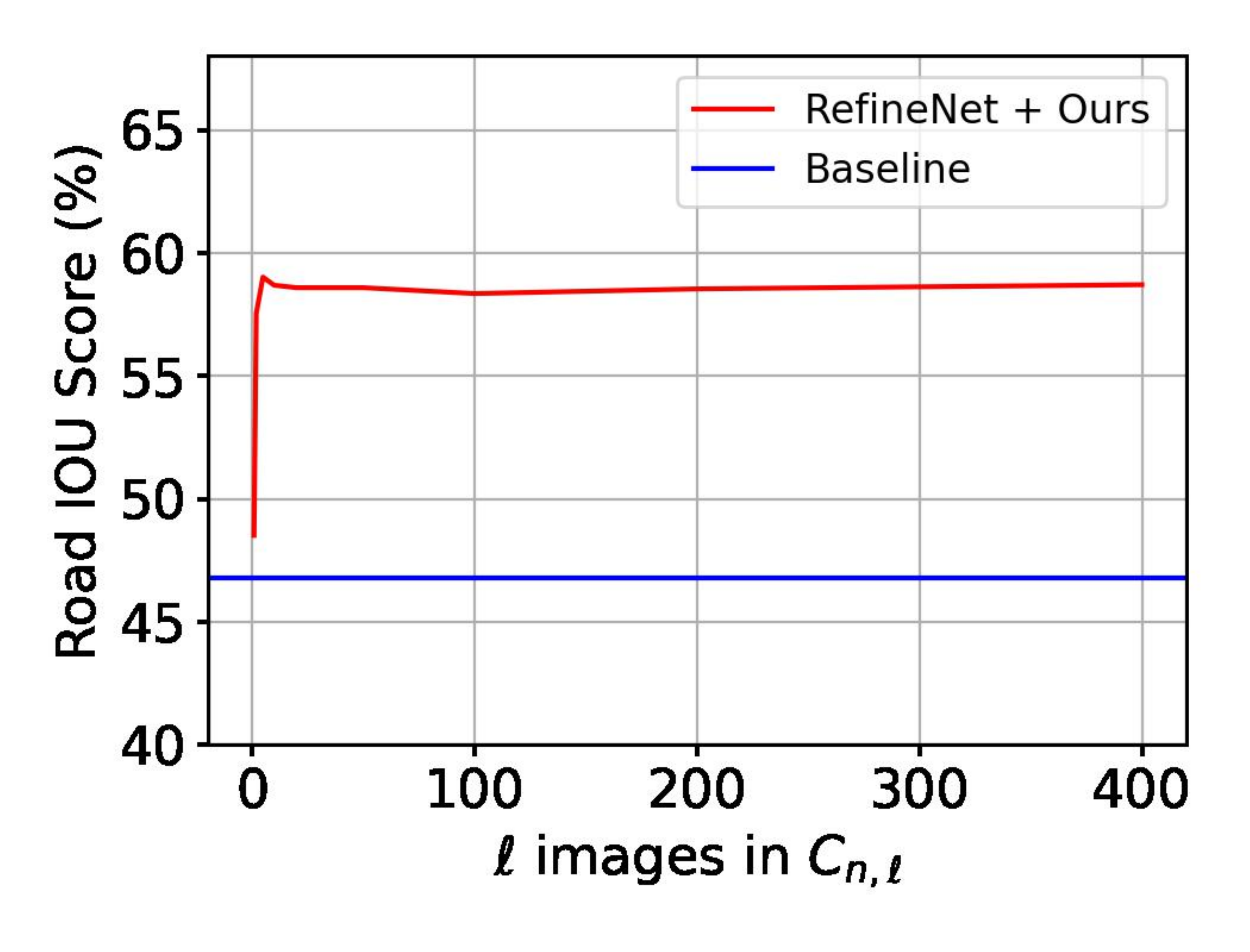} \\
    (a) & (b)
    \end{tabular}
    \caption{Sensitivity analysis. (a) Road IOU varies significantly when generating priors using different VPR networks: NetVLAD (NV) and HybridNet (HN) and mildly over different layers: fully connected (FC7) or convolutional (C6 and C3). (b) Road IOU is stable over a large range when varying top $\ell$ reference candidates for computing $\mathcal{C}_{n,\ell}$ in dynamic weighting scheme, as demonstrated with WildDash Night.}
    \label{fig:DZCoverageAndLayers}
\end{figure}

\subsection{VPR Method Ablations}
\label{sec:VPRAblate}
Unlike normal VPR research, here we focus on retrieving \textit{similar} images which \textit{do not come from the same query place}, but that are structurally similar. Structural similarity is better enforced by VPR methods based on viewpoint-sensitive encoding (e.g. HybridNet~\cite{chen2017deep}), rather than viewpoint-invariant encoding (e.g. NetVLAD~\cite{arandjelovic2016netvlad}). Figure~\ref{fig:DZCoverageAndLayers}(a) shows the effect of choosing a different network or layer for selecting the priors. Using HybridNet (HN) features from different layers -- fully connected (FC7), $conv6$ (C6) and $conv3$ (C3) -- consistently improves road segmentation more than using NetVLAD (NV). $conv6$ (the last convolutional layer) was particularly useful for night-time conditions, likely due to its high receptive field, as compared to $conv3$, and explicit retention of spatial layout, as compared to $FC7$. These findings support previous analysis~\cite{chen2014convolutional,sunderhauf2015performance,garg2018dont}, discussing viewpoint-invariance and the extent of semantic scene understanding across different layers.

\begin{table}
\caption{Change in road IOU when training for binary road segmentation using PyTorch RefineNet~\cite{lengyel2021zero}.}
\centering

\begin{tabular}{@{}lcccccc}
\toprule
\textbf{Dataset} & \multicolumn{1}{c}{Dark Zurich}  & \multicolumn{3}{c}{Wild Dash 2}        &     BDD & All \\

\cmidrule(lr{0.75em}){2-2}
\cmidrule(lr{0.75em}){3-5}
\cmidrule(lr{0.75em}){6-6}
\cmidrule(lr{0.75em}){7-7}
\textbf{Method}  & Night (Test)  & Night       & Snow        & Rain        & Night & Avg.       \\
\midrule
RefineNet        & 1.7          & -3.4         & -14.3           & -10.5          & -3 & -6         \\
\bottomrule
\end{tabular}
\label{tab:RoadOnly}
\end{table}

\subsection{Prior Parameterization Ablations}
In our dynamic weighting scheme, we leverage the top $\ell$ most similar places according to VPR to form an expectation of what the semantic composition ($\mathcal{C}_{n, \ell}$) in a query image should be. To determine the top $\ell$ candidates that would provide sufficient data to compute this, we observed the effect of varying $\ell$ on the road IOU performance. Figure~\ref{fig:DZCoverageAndLayers}(b) demonstrates that the performance of the dynamic weighting scheme is insensitive to the number of candidates used. The system experiences an initial increase in performance when increasing $\ell$, but reaches a plateau and a point where the amount of data is computationally detrimental. We chose to use $\ell$ = 10 as a compromise between the computation time required and the potential for minimal performance increase.

\subsection{Specialised Road Segmentation Networks}
In our investigation we conducted two experiments to compare the performance of our method used with segmentation networks trained for all semantic classes against networks specialised for road segmentation. Table~\ref{tab:singleTable} shows results for the YOLOP network~\cite{wu2021yolop} which was trained on the BDD100k dataset to predict the driveable area in an image. These results show that this specialised network was not able to outperform our method.

In addition, Table~\ref{tab:RoadOnly} provides the performance difference between a RefineNet model trained for all semantic classes and one trained for binary road and non-road segmentation. For these experiments, we use a different PyTorch implementation of RefineNet\footnote{https://github.com/Attila94/CIConv} to the model used in Table~\ref{tab:singleTable} in order to simplify the training process. Despite the 19 and 2-class models achieving 97.3\% and 97.1\% road IOU on the Cityscapes validation set and both achieving 80\% on the test set (using default hyperparameters with $batchsize=4$, $hflip$, $rc$, $jitter=0.3$ and $scale=0.3$), we observed that the 2-class model trained specifically for road segmentation exhibited poor generalisation to unseen domains as compared to the 19-class model.

\subsection{Using Ground Truth Prior Labels}
In some scenarios, ground truth labels would be available for the set of similar place priors. We tested the system across the WildDash 2 datasets using ground truth annotations for the priors, converted to pseudo logit values through adopting the average confidence observed for each class from the base network on daytime images. Table~\ref{tab:GTPriorTable} shows that, if available, using ground truth labels can further improve segmentation results. However, we re-iterate that the performance advantages of the core technique presented in this paper do not require this ground truth.

\begin{table}
\centering
\caption{WildDash 2 RefineNet Road IOU using GT Priors.}
\begin{tabular}{lcc} 
\toprule
Method & Ours without GT &  Ours with GT\\ 
\midrule
WD Night & 58.7 & \textbf{63.4} \\
WD Snow & 70.6 & \textbf{71.2} \\
WD Rain & 70.8 & \textbf{71.2} \\
\bottomrule
\end{tabular}%
\label{tab:GTPriorTable}
\end{table}

\section{Discussion and Future Work}
\label{DISCUSSIONANDFUTURE WORK}

Here we have shown that solving the difficult problem of road segmentation under challenging conditions can be made more tractable by instead solving two related but easier problems: searching for imagery of similar (but geographically distinct) places obtained under ideal conditions, then using the segmentations from those places as a prior to improve performance for the current place. The approach has several practically attractive properties over current techniques and achieves state-of-the-art performance across a range of challenging environmental conditions, and a universal improvement in all tested baseline segmentation methods.

There are many areas of opportunity for future improvements. To aid the search for similar places obtained under \textit{ideal} conditions (because it yields good segmentation results), condition-recognition techniques could be used~\cite{roser2008classification}. While we have re-evaluated the utility of multiple VPR techniques for this novel task, there's much potential for re-examining how to develop and train VPR techniques for this aim. We also intend to further examine the utility of the technique beyond the road class to all types of segmentation classes.

Similar places could also be found by applying VPR on a superpixels or semantically-segmented region basis, reconstructing a similar place in aggregate from multiple matched sub-regions. Finally, the VPR-driven prior identification process proposed here could potentially be expanded to improve a wide range of tasks beyond semantic segmentation that degrade in challenging environmental conditions.

\section*{Acknowledgement}
This research was partially supported by funding from ARC Laureate Fellowship FL210100156 to MM.
This work received funding from the Australian Government, via grant AUSMURIB000001 associated with ONR MURI grant N00014-19-1-2571. The authors acknowledge continued support from the Queensland University of Technology (QUT) through the Centre for Robotics.


\bibliographystyle{IEEEtran}
\bibliography{References}

\begin{thebibliography}{10}
\providecommand{\url}[1]{#1}
\csname url@rmstyle\endcsname
\providecommand{\newblock}{\relax}
\providecommand{\bibinfo}[2]{#2}
\providecommand\BIBentrySTDinterwordspacing{\spaceskip=0pt\relax}
\providecommand\BIBentryALTinterwordstretchfactor{4}
\providecommand\BIBentryALTinterwordspacing{\spaceskip=\fontdimen2\font plus
\BIBentryALTinterwordstretchfactor\fontdimen3\font minus
  \fontdimen4\font\relax}
\providecommand\BIBforeignlanguage[2]{{%
\expandafter\ifx\csname l@#1\endcsname\relax
\typeout{** WARNING: IEEEtran.bst: No hyphenation pattern has been}%
\typeout{** loaded for the language `#1'. Using the pattern for}%
\typeout{** the default language instead.}%
\else
\language=\csname l@#1\endcsname
\fi
#2}}

\bibitem{garg2020semantics}
\BIBentryALTinterwordspacing
S.~Garg, N.~S{\"u}nderhauf, F.~Dayoub, D.~Morrison, A.~Cosgun, G.~Carneiro,
  Q.~Wu, T.-J. Chin, I.~Reid, S.~Gould, P.~Corke, and M.~Milford, ``Semantics
  for robotic mapping, perception and interaction: A survey,'' \emph{Found.
  Trends Robot.}, vol.~8, no. 1–2, pp. 1--224, 2020. [Online]. Available:
  \url{http://dx.doi.org/10.1561/2300000059}
\BIBentrySTDinterwordspacing

\bibitem{chen2018encoder}
L.-C. Chen, Y.~Zhu, G.~Papandreou, F.~Schroff, and H.~Adam, ``Encoder-decoder
  with atrous separable convolution for semantic image segmentation,'' in
  \emph{Proceedings of the European conference on computer vision (ECCV)},
  2018, pp. 801--818.

\bibitem{cordts2016cityscapes}
M.~Cordts, M.~Omran, S.~Ramos, T.~Rehfeld, M.~Enzweiler, R.~Benenson,
  U.~Franke, S.~Roth, and B.~Schiele, ``The cityscapes dataset for semantic
  urban scene understanding,'' in \emph{Proceedings of the IEEE Conference on
  Computer Vision and Pattern Recognition}, 2016, pp. 3213--3223.

\bibitem{sakaridis2020map}
C.~Sakaridis, D.~Dai, and L.~Van~Gool, ``Map-guided curriculum domain
  adaptation and uncertainty-aware evaluation for semantic nighttime image
  segmentation,'' \emph{IEEE Transactions on Pattern Analysis and Machine
  Intelligence}, 2020.

\bibitem{dai2018dark}
D.~Dai and L.~Van~Gool, ``Dark model adaptation: Semantic image segmentation
  from daytime to nighttime,'' in \emph{2018 21st International Conference on
  Intelligent Transportation Systems (ITSC)}.\hskip 1em plus 0.5em minus
  0.4em\relax IEEE, 2018, pp. 3819--3824.

\bibitem{porav2019don}
H.~Porav, T.~Bruls, and P.~Newman, ``Don’t worry about the weather:
  Unsupervised condition-dependent domain adaptation,'' in \emph{2019 IEEE
  Intelligent Transportation Systems Conference (ITSC)}.\hskip 1em plus 0.5em
  minus 0.4em\relax IEEE, 2019, pp. 33--40.

\bibitem{long2015fully}
J.~Long, E.~Shelhamer, and T.~Darrell, ``Fully convolutional networks for
  semantic segmentation,'' in \emph{Proceedings of the IEEE conference on
  computer vision and pattern recognition}, 2015, pp. 3431--3440.

\bibitem{girshick2015region}
R.~Girshick, J.~Donahue, T.~Darrell, and J.~Malik, ``Region-based convolutional
  networks for accurate object detection and segmentation,'' \emph{IEEE
  transactions on pattern analysis and machine intelligence}, vol.~38, no.~1,
  pp. 142--158, 2015.

\bibitem{badrinarayanan2017segnet}
V.~Badrinarayanan, A.~Kendall, and R.~Cipolla, ``Segnet: A deep convolutional
  encoder-decoder architecture for image segmentation,'' \emph{IEEE
  transactions on pattern analysis and machine intelligence}, vol.~39, no.~12,
  pp. 2481--2495, 2017.

\bibitem{zhao2017pyramid}
H.~Zhao, J.~Shi, X.~Qi, X.~Wang, and J.~Jia, ``Pyramid scene parsing network,''
  in \emph{Proceedings of the IEEE conference on computer vision and pattern
  recognition}, 2017, pp. 2881--2890.

\bibitem{tao2020hierarchical}
A.~Tao, K.~Sapra, and B.~Catanzaro, ``Hierarchical multi-scale attention for
  semantic segmentation,'' \emph{arXiv preprint arXiv:2005.10821}, 2020.

\bibitem{lei2020semantic}
Y.~Lei, T.~Emaru, A.~A. Ravankar, Y.~Kobayashi, and S.~Wang, ``Semantic image
  segmentation on snow driving scenarios,'' in \emph{2020 IEEE International
  Conference on Mechatronics and Automation (ICMA)}.\hskip 1em plus 0.5em minus
  0.4em\relax IEEE, 2020, pp. 1094--1100.

\bibitem{vachmanus2020semantic}
S.~Vachmanus, A.~A. Ravankar, T.~Emaru, and Y.~Kobayashi, ``Semantic
  segmentation for road surface detection in snowy environment,'' in \emph{2020
  59th Annual Conference of the Society of Instrument and Control Engineers of
  Japan (SICE)}.\hskip 1em plus 0.5em minus 0.4em\relax IEEE, 2020, pp.
  1381--1386.

\bibitem{li2020segmenting}
C.~Li, W.~Xia, Y.~Yan, B.~Luo, and J.~Tang, ``Segmenting objects in day and
  night: Edge-conditioned cnn for thermal image semantic segmentation,''
  \emph{IEEE Transactions on Neural Networks and Learning Systems}, 2020.

\bibitem{valada2017adapnet}
A.~Valada, J.~Vertens, A.~Dhall, and W.~Burgard, ``Adapnet: Adaptive semantic
  segmentation in adverse environmental conditions,'' in \emph{2017 IEEE
  International Conference on Robotics and Automation (ICRA)}.\hskip 1em plus
  0.5em minus 0.4em\relax IEEE, 2017, pp. 4644--4651.

\bibitem{pfeuffer2020robust}
A.~Pfeuffer and K.~Dietmayer, ``Robust semantic segmentation in adverse weather
  conditions by means of fast video-sequence segmentation,'' in \emph{2020 IEEE
  23rd International Conference on Intelligent Transportation Systems
  (ITSC)}.\hskip 1em plus 0.5em minus 0.4em\relax IEEE, 2020, pp. 1--6.

\bibitem{pfeuffer2019robust}
------, ``Robust semantic segmentation in adverse weather conditions by means
  of sensor data fusion,'' in \emph{2019 22th International Conference on
  Information Fusion (FUSION)}.\hskip 1em plus 0.5em minus 0.4em\relax IEEE,
  2019, pp. 1--8.

\bibitem{sun2019see}
L.~Sun, K.~Wang, K.~Yang, and K.~Xiang, ``See clearer at night: towards robust
  nighttime semantic segmentation through day-night image conversion,'' in
  \emph{Artificial Intelligence and Machine Learning in Defense Applications},
  vol. 11169.\hskip 1em plus 0.5em minus 0.4em\relax International Society for
  Optics and Photonics, 2019, p. 111690A.

\bibitem{romera2019bridging}
E.~Romera, L.~M. Bergasa, K.~Yang, J.~M. Alvarez, and R.~Barea, ``Bridging the
  day and night domain gap for semantic segmentation,'' in \emph{2019 IEEE
  Intelligent Vehicles Symposium (IV)}.\hskip 1em plus 0.5em minus 0.4em\relax
  IEEE, 2019, pp. 1312--1318.

\bibitem{goodfellow2014generative}
I.~Goodfellow, J.~Pouget-Abadie, M.~Mirza, B.~Xu, D.~Warde-Farley, S.~Ozair,
  A.~Courville, and Y.~Bengio, ``Generative adversarial nets,'' in
  \emph{Advances in neural information processing systems}, 2014, pp.
  2672--2680.

\bibitem{zhu2017unpaired}
J.-Y. Zhu, T.~Park, P.~Isola, and A.~A. Efros, ``Unpaired image-to-image
  translation using cycle-consistent adversarial networks,'' in
  \emph{Proceedings of the IEEE international conference on computer vision},
  2017, pp. 2223--2232.

\bibitem{jonchery2018stixel}
S.~Jonchery, G.~Bresson, B.~Vallet, and R.~{\.Z}bikowski, ``A stixel approach
  for enhancing semantic image segmentation using prior map information,'' in
  \emph{2018 15th International Conference on Control, Automation, Robotics and
  Vision (ICARCV)}.\hskip 1em plus 0.5em minus 0.4em\relax IEEE, 2018, pp.
  1715--1720.

\bibitem{schroeder2019using}
B.~Schroeder and A.~Alahi, ``Using a priori knowledge to improve scene
  understanding,'' in \emph{Proceedings of the IEEE Conference on Computer
  Vision and Pattern Recognition Workshops}, 2019, pp. 487--489.

\bibitem{lowry2016visual}
S.~Lowry, N.~S{\"u}nderhauf, P.~Newman, J.~J. Leonard, D.~Cox, P.~Corke, and
  M.~J. Milford, ``Visual place recognition: A survey,'' \emph{IEEE
  Transactions on Robotics}, vol.~32, no.~1, pp. 1--19, 2016.

\bibitem{cummins2008fab}
M.~Cummins and P.~Newman, ``Fab-map: Probabilistic localization and mapping in
  the space of appearance,'' \emph{The International Journal of Robotics
  Research}, vol.~27, no.~6, pp. 647--665, 2008.

\bibitem{milford2012visual}
M.~Milford, ``Visual route recognition with a handful of bits,'' \emph{Proc.
  2012 Robotics: Science and Systems VIII}, pp. 297--304, 2012.

\bibitem{arandjelovic2016netvlad}
R.~Arandjelovic, P.~Gronat, A.~Torii, T.~Pajdla, and J.~Sivic, ``Netvlad: Cnn
  architecture for weakly supervised place recognition,'' in \emph{Proceedings
  of the IEEE Conference on Computer Vision and Pattern Recognition}, 2016, pp.
  5297--5307.

\bibitem{chen2017deep}
Z.~Chen, A.~Jacobson, N.~S{\"u}nderhauf, B.~Upcroft, L.~Liu, C.~Shen, I.~Reid,
  and M.~Milford, ``Deep learning features at scale for visual place
  recognition,'' in \emph{Robotics and Automation (ICRA), 2017 IEEE
  International Conference on}.\hskip 1em plus 0.5em minus 0.4em\relax IEEE,
  2017, pp. 3223--3230.

\bibitem{naseer2017semantics}
T.~Naseer, G.~L. Oliveira, T.~Brox, and W.~Burgard, ``Semantics-aware visual
  localization under challenging perceptual conditions,'' in \emph{IEEE
  International Conference on Robotics and Automation (ICRA)}, 2017.

\bibitem{garg19Semantic}
S.~Garg, N.~S{\"u}nderhauf, and M.~Milford, ``Semantic-geometric visual place
  recognition: A new perspective for reconciling opposing views,'' \emph{The
  International Journal of Robotics Research}, 2019.

\bibitem{hendrycks17baseline}
D.~Hendrycks and K.~Gimpel, ``A baseline for detecting misclassified and
  out-of-distribution examples in neural networks,'' \emph{Proceedings of
  International Conference on Learning Representations}, 2017.

\bibitem{xia2020synthesize}
Y.~Xia, Y.~Zhang, F.~Liu, W.~Shen, and A.~Yuille, ``Synthesize then compare:
  Detecting failures and anomalies for semantic segmentation,'' in
  \emph{Computer Vision -- ECCV 2020}.\hskip 1em plus 0.5em minus 0.4em\relax
  Springer International Publishing, 2020, pp. 145--161.

\bibitem{SDV19}
C.~Sakaridis, D.~Dai, and L.~Van~Gool, ``Guided curriculum model adaptation and
  uncertainty-aware evaluation for semantic nighttime image segmentation,'' in
  \emph{The IEEE International Conference on Computer Vision (ICCV)}, 2019.

\bibitem{Zendel_2018_ECCV}
O.~Zendel, K.~Honauer, M.~Murschitz, D.~Steininger, and G.~F. Dominguez,
  ``Wilddash - creating hazard-aware benchmarks,'' in \emph{Proceedings of the
  European Conference on Computer Vision (ECCV)}, September 2018.

\bibitem{bdd100k}
F.~Yu, H.~Chen, X.~Wang, W.~Xian, Y.~Chen, F.~Liu, V.~Madhavan, and T.~Darrell,
  ``Bdd100k: A diverse driving dataset for heterogeneous multitask learning,''
  in \emph{The IEEE Conference on Computer Vision and Pattern Recognition
  (CVPR)}, June 2020.

\bibitem{Lin:2017:RefineNet}
G.~Lin, A.~Milan, C.~Shen, and I.~Reid, ``Refine{N}et: {M}ulti-path refinement
  networks for high-resolution semantic segmentation,'' in \emph{CVPR}, July
  2017.

\bibitem{tsai2018learning}
Y.-H. Tsai, W.-C. Hung, S.~Schulter, K.~Sohn, M.-H. Yang, and M.~Chandraker,
  ``Learning to adapt structured output space for semantic segmentation,'' in
  \emph{Proceedings of the IEEE conference on computer vision and pattern
  recognition}, 2018, pp. 7472--7481.

\bibitem{vu2019advent}
T.-H. Vu, H.~Jain, M.~Bucher, M.~Cord, and P.~P{\'e}rez, ``Advent: Adversarial
  entropy minimization for domain adaptation in semantic segmentation,'' in
  \emph{Proceedings of the IEEE/CVF Conference on Computer Vision and Pattern
  Recognition}, 2019, pp. 2517--2526.

\bibitem{li2019bidirectional}
Y.~Li, L.~Yuan, and N.~Vasconcelos, ``Bidirectional learning for domain
  adaptation of semantic segmentation,'' in \emph{Proceedings of the IEEE/CVF
  Conference on Computer Vision and Pattern Recognition}, 2019, pp. 6936--6945.

\bibitem{chen2014convolutional}
Z.~Chen, O.~Lam, A.~Jacobson, and M.~Milford, ``Convolutional neural
  network-based place recognition,'' in \emph{Australasian Conference on
  Robotics and Automation}, vol.~2, 2014, p.~4.

\bibitem{sunderhauf2015performance}
N.~S{\"u}nderhauf, S.~Shirazi, F.~Dayoub, B.~Upcroft, and M.~Milford, ``On the
  performance of convnet features for place recognition,'' in \emph{Intelligent
  Robots and Systems (IROS), 2015 IEEE/RSJ International Conference on}.\hskip
  1em plus 0.5em minus 0.4em\relax IEEE, 2015, pp. 4297--4304.

\bibitem{garg2018dont}
S.~Garg, N.~Suenderhauf, and M.~Milford, ``Don't look back: Robustifying place
  categorization for viewpoint- and condition-invariant place recognition,'' in
  \emph{IEEE International Conference on Robotics and Automation (ICRA)}, 2018.

\bibitem{lengyel2021zero}
A.~Lengyel, S.~Garg, M.~Milford, and J.~C. van Gemert, ``Zero-shot domain
  adaptation with a physics prior,'' in \emph{The IEEE International Conference
  on Computer Vision (ICCV)}, 2021.

\bibitem{wu2021yolop}
D.~Wu, M.~Liao, W.~Zhang, and X.~Wang, ``Yolop: You only look once for panoptic
  driving perception,'' \emph{arXiv preprint arXiv:2108.11250}, 2021.

\bibitem{roser2008classification}
M.~Roser and F.~Moosmann, ``Classification of weather situations on single
  color images,'' in \emph{2008 IEEE Intelligent Vehicles Symposium}.\hskip 1em
  plus 0.5em minus 0.4em\relax IEEE, 2008, pp. 798--803.

\end{thebibliography}

\end{document}